\pdfoutput=1

\documentclass[11pt]{article}

\usepackage[table]{xcolor}
\usepackage[]{acl}

\usepackage{times}
\usepackage{latexsym}

\usepackage[T1]{fontenc}

\usepackage[utf8]{inputenc}

\usepackage{microtype}

\usepackage{inconsolata}

\usepackage{graphicx}

\usepackage{amsmath}
\usepackage{amsthm}
\usepackage{amsfonts}
\usepackage{booktabs}
\usepackage{multirow}
\usepackage{xspace}
\usepackage{xcolor}
\usepackage{tcolorbox}
\usepackage{xargs}
\usepackage{MnSymbol,bbding,pifont}
\usepackage{colortbl}
\usepackage{subcaption}
\usepackage{graphicx}
\usepackage{tikz}
\usepackage{float}

\usepackage{fontawesome5}
\usepackage{soul}

\definecolor{oc-gray-0}{HTML}{F8F9FA}
\definecolor{oc-gray-1}{HTML}{F1F3F5}
\definecolor{oc-gray-2}{HTML}{E9ECEF}
\definecolor{oc-gray-3}{HTML}{DEE2E6}
\definecolor{oc-gray-4}{HTML}{CED4DA}
\definecolor{oc-gray-5}{HTML}{ADB5BD}
\definecolor{oc-gray-6}{HTML}{868E96}
\definecolor{oc-gray-7}{HTML}{495057}
\definecolor{oc-gray-8}{HTML}{343A40}
\definecolor{oc-gray-9}{HTML}{212529}
\definecolor{oc-black}{HTML}{000000}
\definecolor{oc-blue-0}{HTML}{E7F5FF}
\definecolor{oc-blue-1}{HTML}{d0ebff}
\definecolor{oc-blue-7}{HTML}{1C7ED6}
\definecolor{oc-blue-8}{HTML}{1971C2}
\definecolor{oc-blue-9}{HTML}{1864AB}
\definecolor{oc-lime-0}{HTML}{F4FCE3}
\definecolor{oc-lime-8}{HTML}{66A80F}
\definecolor{oc-lime-9}{HTML}{5C940D}
\definecolor{oc-green-0}{HTML}{EBFBEE}
\definecolor{oc-green-8}{HTML}{2F9E44}
\definecolor{oc-green-9}{HTML}{2B8A3E}
\definecolor{oc-orange-0}{HTML}{FFF4E6}
\definecolor{oc-orange-8}{HTML}{E8590C}
\definecolor{oc-orange-9}{HTML}{D9480F}
\definecolor{oc-maroon}{HTML}{A61E4D}

\definecolor{user-yellow}{HTML}{FFD43B}
\definecolor{oc-maroon}{HTML}{A61E4D}
\newcommand{\tulu}{\textsc{T\"ulu}\xspace}
\newcommand{\llama}{llama\xspace}
\newcommand{\tulumix}{\textsc{T\"ulu V2 Mix}\xspace}
\newcommand{\red}[1]{\textcolor{red}{#1}}

\newcommand{\lime}[1]{\textcolor{oc-green-8}{#1}}

\newtcolorbox{promptbox}[2][Prompt]{
colback=black!4!white, 
arc=5pt, 
boxrule=1.1pt,
fonttitle=\bfseries,
title=#1, 
before upper={\small}, 
fontupper=\selectfont\footnotesize, 
colframe=#2, 
}

\iffalse   
\newcommand{\yl}[1]{}
\else
\newcommand{\yl}[1]{\textcolor{cyan}{\textbf{YL:} #1}}
\fi

\newcommand{\llmbarnatural}{\texttt{LLMBar-Natural}\xspace}
\newcommand{\adversarial}{\texttt{LLMBar-Adversarial}\xspace}
\newcommand{\mtbench}{\texttt{MTBench}\xspace}
\newcommand{\instrusum}{\texttt{InstruSum}\xspace}

\newcommand{\natshort}{\texttt{Nat.}\xspace}
\newcommand{\advshort}{\texttt{Adv.}\xspace}
\newcommand{\mtshort}{\texttt{MT.}\xspace}
\newcommand{\insshort}{\texttt{Ins.}\xspace}

\newcommand{\bettersig}{$^\uparrow$}
\newcommand{\worsesig}{$^\downarrow$}
\newcommand{\cparagraph}[1]{\noindent\textbf{#1}}

\newcommand{\ours}{\textsc{ReIFE}\xspace}

\title{\ours: Re-evaluating Instruction-Following Evaluation}

\author{
 Yixin Liu\Thanks{~Equal contribution}$^{1}$ 
 \quad \textbf{Kejian Shi}\footnotemark[1]$^{1}$
 \quad \textbf{Alexander R. Fabbri}$^{2}$
 \quad \textbf{Yilun Zhao}$^{1}$ \\
 \quad \textbf{Peifeng Wang}$^{2}$ 
 \quad \textbf{Chien-Sheng Wu}$^{2}$ 
  \quad \textbf{Shafiq Joty}$^{2}$ 
 \quad \textbf{Arman Cohan}$^{1,3}$ \vspace{6pt}\\
  $^1$Yale University\quad 
  $^2$Salesforce AI\quad
  $^3$Allen Institute for AI
  \vspace{6pt}\\
  \texttt{yixin.liu@yale.edu, kejian.shi@yale.edu, arman.cohan@yale.edu}
 }
 
\begin{document}
\maketitle
\begin{abstract}
The automatic evaluation of instruction following typically involves using large language models (LLMs) to assess response quality. 
However, there is a lack of comprehensive evaluation of these LLM-based evaluators across two dimensions: the base LLMs and the evaluation protocols.
Therefore, we present a thorough meta-evaluation of instruction following, including 25 base LLMs and 15 recently proposed evaluation protocols, on 4 human-annotated datasets, assessing the evaluation accuracy of the LLM-evaluators.
Our evaluation allows us to identify the best-performing base LLMs and evaluation protocols with a high degree of robustness.
Moreover, our large-scale evaluation reveals:
(1) Base LLM performance ranking remains largely consistent across evaluation protocols, with less capable LLMs showing greater improvement from protocol enhancements;
(2) Robust evaluation of evaluation protocols requires many base LLMs with varying capability levels, as protocol effectiveness can depend on the base LLM used;
(3) Evaluation results on different datasets are not always consistent, so a rigorous evaluation requires multiple datasets with distinctive features.
We release our meta-evaluation suite \ours,\footnote{\ours stands for \textbf{R}e-\textbf{e}valuation of \textbf{I}nstruction-\textbf{F}ollowing \textbf{E}valuation: \url{https://github.com/yale-nlp/ReIFE}.} which provides the codebase and evaluation result collection for more than 500 LLM-evaluator configurations, to support future research in instruction-following evaluation.
\end{abstract}

\section{Introduction}

\begin{figure*}[t!]
    \centering
    \includegraphics[width=0.9\linewidth]{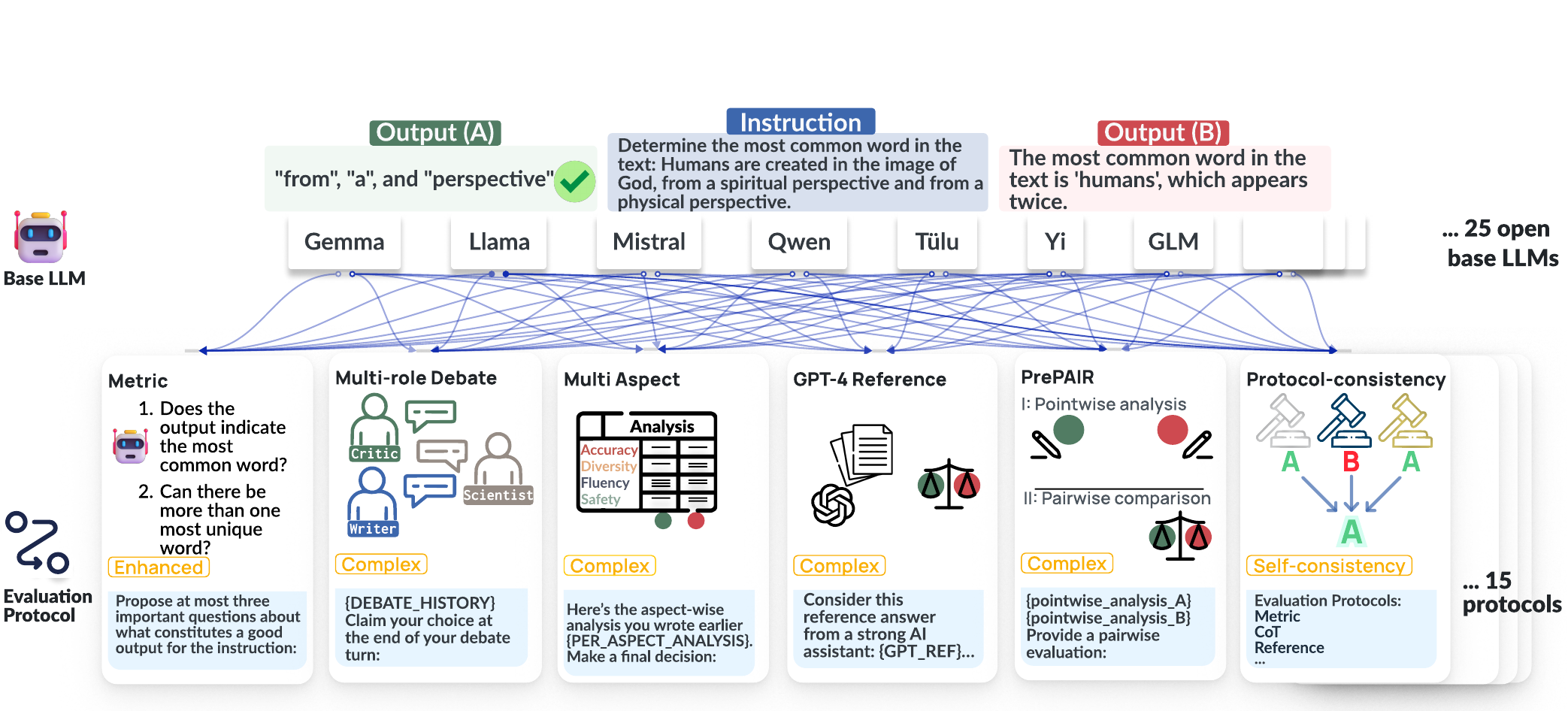}
    \caption{\label{fig:intro}Overview of our large-scale meta-evaluation study of instruction-following evaluation.
    We evaluate the capabilities of 25 open-source base LLMs and 15 evaluation protocols, resulting in a total of 375 LLM-evaluators --  evaluation methods that perform the evaluations using the base LLMs by following the evaluation protocols.
    }
\end{figure*}

The ability to follow human instructions has become an important evaluation aspect for large language models (LLMs), indicating their alignment with human users~\cite{NEURIPS2022_b1efde53}.
Recently, due to their better correlation with human judgments compared with traditional evaluation metrics, the LLMs themselves are often used as judges of the model output quality for generative tasks including instruction following~\cite{liu-etal-2023-g, fu2023gptscore, zheng2024judging}.
These LLM-based evaluation methods are an essential component of the most widely used automatic benchmarks for instruction-following evaluation, such as AlpacaEval~\cite{alpaca_eval} and MTBench~\cite{zheng2024judging}, where a strong LLM is used to evaluate the quality of model responses.
Moreover, they can be used as reward models for instruction fine-tuning of LLMs in both distillation and self-improvement settings~\citep{tunstall2023zephyr, yuan2024selfrewarding}.
However, recent studies have identified various limitations of LLM-based evaluation methods, including low self-consistency rates in their predictions, positional biases, and a preference for their own outputs~\cite{liu-etal-2023-g,wang-etal-2024-large-language-models-fair, zheng2024judging, panickssery2024llm}.

Therefore, the evaluation of LLM-based evaluations is critically important.
Such evaluations of evaluation methods, or meta-evaluation, usually involve comparing the automatic evaluation results against human evaluation~\cite{liu-etal-2023-g, dubois2024alpacafarm, zeng2024evaluating}.
These evaluations of LLM-evaluators assess two dimensions: (1) the capabilities of \textit{base LLMs} in performing the evaluation task and (2) the effectiveness of \textit{evaluation protocols} — the methods by which base LLMs are used to perform evaluation, e.g., pairwise comparison as in AlpacaEval or pointwise scoring as in MTBench.\footnote{We use ``LLM-evaluator'' to refer to an evaluation method that combines a base LLM and an evaluation protocol.}
Existing work~\cite{zheng2024judging, wang-etal-2024-large-language-models-fair, zeng2024evaluating} often lacks comprehensiveness in one or both of these dimensions, and more thorough evaluations are needed.

We argue that the following two directions are crucial for a more comprehensive, rigorous evaluation of LLM-evaluators for instruction following: 
(1) \textbf{Including the diverse set of base LLMs} for the evaluation of evaluation protocols -- while various evaluation protocols have been proposed recently~\cite{gong2023coascore, saha2023branchsolvemerge, chan2024chateval, jeong2024prepair}, meta-evaluation studies of these evaluation protocols often lack scale in the number of LLMs used. 
For example, LLMBar~\citep{zeng2024evaluating} uses only 5 LLMs to compare different evaluation protocols. 
As a result, it remains unclear whether the improvements observed in recently introduced evaluation protocols are robust and generalizable across base LLMs with varying performance levels.
Therefore, we aim to conduct an evaluation with a larger and more diverse set of base LLMs to ensure a more rigorous examination of the evaluation protocols.
(2) \textbf{Expanding the pool of evaluation protocols} for the evaluation of base LLMs -- various related studies use only a limited number of evaluation protocols when assessing the evaluation capabilities of different base LLMs~\cite{liu-etal-2023-g, dubois2024alpacafarm}.
However, LLMs' performance can be sensitive to prompt design~\cite{sclar2024quantifying}, raising doubts about the reliability of using a single protocol for evaluation.
Consequently, we aim to achieve a more reliable evaluation of base LLMs by including a larger set of recently proposed evaluation protocols to account for performance variations from prompt/protocol configurations.

Based on these goals, we present an in-depth meta-evaluation with the following components:

\noindent (1) We perform a robust baseline evaluation across 4 meta-evaluation datasets by evaluating 38 base LLMs and 3 evaluation protocols used in existing benchmarks: Alpacaeval, ArenaHard~\cite{li2024crowdsourced}, and WildBench~\cite{lin2024wildbench} (\S\ref{sec:status-quo}).

\noindent (2) We gather 15 evaluation protocols based on previous work, applying a unified prompting style for a fair comparison, and evaluate their average performance with 25 open-source LLMs (\S\ref{sec:protocols}).

\noindent (3) We leverage the large number of 375 LLM-evaluators evaluated in \S\ref{sec:protocols} to perform an in-depth analysis of the practice of meta-evaluation itself, addressing research questions on base LLMs, evaluation protocols, and datasets used in the meta-evaluation process (\S\ref{sec:analysis}).

Our large-scale meta-evaluation, as outlined in Figure~\ref{fig:intro}, enables a thorough examination of the current progress in LLM-based instruction-following evaluation, providing a solid foundation for developing evaluation protocols and evaluating base LLMs' evaluation capabilities.
We make our meta-evaluation suite \ours publicly available, which contains the codebase and evaluation result collection for more than 500 LLM-evaluators we evaluated,  and we summarize our key findings below:

\paragraph{Findings}
\noindent (1) When used in conjunction with 15 evaluation protocols, Llama-3.1-405B~\cite{dubey2024llama} is the best open-source base LLM we evaluated (Table~\ref{tab:models}), which approaches state-of-the-art proprietary LLM performance (Table~\ref{tab:baseline}).

\noindent (2) The evaluation protocols used in 3 widely used benchmarks fail to outperform even the base evaluation protocol evaluated in this work (Table~\ref{tab:benchmark-protocol}).
In contrast, the recently introduced evaluation protocol, \texttt{prepair}~\citep{jeong2024prepair}, achieves the highest average performance across 25 open-source LLMs, with 7 of the protocols evaluated significantly outperforming the base protocol (Table~\ref{tab:protocol-avg}).

\noindent (3) 
The performance ranking of different base LLMs is largely consistent across different evaluation protocols, suggesting that evaluating different base LLMs' evaluation capabilities with a single evaluation protocol is likely to yield reliable results (\S\ref{subsec:llm-analysis}).
However, the benefits of advanced protocols vary across LLMs, with less capable LLMs more likely to gain greater improvements (Table~\ref{tab:models-improvement}).

\noindent (4) The effectiveness of evaluation protocols depends significantly on the base LLMs used (\S\ref{subsec:analysis-protocols}). 
For example, although \texttt{prepair} achieves the highest average performance, it ranks only seventh among the 15 evaluation protocols when comparing their optimal performance achieved with the most compatible base LLMs (Table~\ref{tab:protocol-best}). 
This highlights the need to use multiple base LLMs with varying performance levels for reliable evaluation of evaluation protocols.

\noindent (5) Different meta-evaluation datasets can exhibit varying difficulty levels, and the LLM-evaluator rankings on these datasets do not always show a strong positive correlation, demonstrating the importance of incorporating diverse datasets for a more comprehensive meta-evaluation (\S\ref{subsec:analysis-dataset}).

\section{Related Work}
\label{sec:related-work}

\paragraph{LLM-based Evaluation} Using LLMs as evaluators has become a promising approach for assessing text generation quality~\citep{chiang-lee-2023-large, fu2023gptscore, liu-etal-2023-g} in tasks like summarization \citep{fu2023gptscore, liu-etal-2023-g, liu-etal-2023-revisiting} and instruction-following \citep{zheng2024judging, zeng2024evaluating, alpaca_eval}. 

Recent work has proposed various advanced LLM-based evaluation methods. 
For example, fine-grained or decomposition-based approaches, such as Chain-of-Aspects \citep{gong2023coascore} and Branch-Solve-Merge \citep{saha2023branchsolvemerge}, can guide LLMs to perform structured analysis by identifying fine-grained differences and providing detailed rationales. 
Agent-based methods, like PRD \citep{li2023prd} and ChatEval \citep{chan2024chateval}, employ multi-role debate settings to bring diverse perspectives to the evaluation process.
Other techniques include probability-weighted scoring \citep{liu-etal-2023-g}, reference-based evaluation \citep{zeng2024evaluating}, and self-consistency decoding \citep{wang2023selfconsistency}. 
Our study investigates the effectiveness of these advanced evaluation protocols on a larger scale, assessing their performance across multiple datasets and base LLMs.

Related studies have also explored fine-tuning LLMs as evaluators for various evaluation tasks including instruction-following evaluation~\cite{li2023generative, wang2024direct}, such as Prometheus~\cite{kim2024prometheus}. 
However, we choose to exclude them from the majority of our evaluation since our focus is on generic LLMs with various evaluation protocols, while the fine-tuned LLMs usually require a fixed evaluation protocol.

\paragraph{Human Evaluation and Meta-Evaluation of Instruction-Following} 

A series of recent studies have conducted human evaluations on instruction-following and/or performed evaluations of automatic evaluators using the collected human annotations~\cite{ zhang2023wider, wang-etal-2024-large-language-models-fair, wang2024pandalm, lan2024criticbench}.
Among them, the annotations from AlpacaFarm~\cite{dubois2024alpacafarm} and MTBench~\citep{zheng2024judging} have become important testbeds for evaluating widely used LLM evaluators. 
\citet{zeng2024evaluating} introduce LLMBar, which consists of high-quality human annotations with a high level of inter-annotator agreement rate. 
RewardBench~\citep{lambert2024rewardbench} provides a benchmark for evaluating reward models used for learning from human or LLM feedback~\citep{NEURIPS2022_b1efde53,bai2022constitutional,tunstall2023zephyr}.
While sharing a similar task format, our evaluation focus is different from theirs because we aim to assess the evaluation capability of generic LLMs instead of dedicated reward models.

\section{Evaluation Settings of \ours}
\label{sec:data_models}
In \ours, we evaluate LLM-based instruction-following evaluations along two dimensions: base LLMs and evaluation protocols (Figure~\ref{fig:intro}), using human evaluations as the gold standard.
Below, we outline the settings of this evaluation.

\paragraph{Datasets}
We use four datasets to evaluate the LLMs' capability of instruction-following.
Each dataset includes human annotations for pairwise comparisons of two outputs from an instruction, with a binary label indicating which output is better in instruction following.
Table~\ref{tab:dataset-info} summarizes the dataset information.
\llmbarnatural and \adversarial are from \citet{zeng2024evaluating}, consisting of data examples examined and edited by the paper authors.
\mtbench~\citep{zheng2024judging} contains expert human annotations made by graduate students for multi-turn conversations. 
\instrusum~\cite{liu2024benchmarking} contains human annotations for instruction-controllable summarization, where the input includes a source article and a specific summary requirement, as a complex instruction.
We only use its annotation data samples with perfect annotator agreement to reduce the annotation noise.
In Appendix~\ref{appx_data_examples}, we show data examples from the datasets.
We selected these datasets due to their varying difficulties, instruction complexity, and human annotation noise.
For example, \instrusum contains much longer instructions than the other datasets, while \mtbench has lower agreement than the others.

\cparagraph{LLM-Evaluator Settings}
Since all the datasets we use contain \textit{pairwise} human evaluations, we evaluate LLM-evaluators under the same pairwise comparison setting for evaluation target alignment.
We use the term ``LLM-evaluator'' to refer to a combination of an \textit{base LLM} and an \textit{evaluation protocol}.
An evaluation protocol defines how the base LLM performs the evaluation, typically using one or more prompts to query it.
By default, we use greedy decoding to ensure deterministic behavior.

\cparagraph{Evaluation Metrics}
We mainly use \textit{\textbf{evaluation accuracy}} to evaluate the LLM-evaluators, which measures the alignment with human evaluations using human annotations as the gold standard.
Since the LLM-evaluators perform pairwise comparisons, to account for potential \textit{position biases}, where the LLM-evaluators may favor either the first or the second output \cite{wang-etal-2024-large-language-models-fair}, we report the averaged evaluation accuracy across two directions, swapping the order of the two outputs.
An auxiliary metric we used is the \textit{\textbf{self-agreement rate}} of the LLM-evaluators in their predictions across two directions in Krippendorff’s alpha~\cite{Krippendorff2011ComputingKA}, measuring their positional biases.
\renewcommand{\arraystretch}{1.1} 
\begin{table}[t!]
\small
\centering
\addtolength{\tabcolsep}{-4pt} 
\begin{tabular}{lrrrrr}
\toprule
 & \textbf{Abbr.} & \textbf{In.L.} & \textbf{Out.L.}  & \textbf{Num.} & \textbf{Agr.}\\
\midrule
\llmbarnatural & \natshort & 53.3 & 56.9 & 100 & 90\% \\
\adversarial & \advshort & 26.7 & 112.4 & 319 & 95\% \\
\mtbench & \mtshort & 58.8 & 192.4 & 200 & 81\% \\ 
\instrusum & \insshort & 1149.8 & 109.2 & 411 & 100\% \\ 
\bottomrule
\end{tabular}
\addtolength{\tabcolsep}{+4pt} 
\caption{Dataset information including abbreviation used (\textbf{Abbr.}), average instruction (\textbf{In.L.}) and output length (\textbf{Out.L.}) in words, number of examples (\textbf{Num.}), and annotation agreement rate (\textbf{Agr.}).
}
\label{tab:dataset-info} 
\end{table}

\section{Baselines}
\label{sec:status-quo}

We first establish baselines for base LLMs and evaluation protocols for evaluating instruction-following for our further investigations.

\subsection{Baselines for Base LLMs}
\label{subsec:llm-baseline}
To benchmark the baseline performance of base LLMs at instruction-following evaluation, we evaluate them with a simple evaluation protocol to construct the corresponding LLM-evaluators.
This \textit{base} evaluation protocol, proposed in \citet{zeng2024evaluating}, requires the LLM-evaluators to directly predict which output is better, with rules to constrain output formats and to avoid potential biases.\footnote{The prompt template is in Appendix~\ref{appx_base_prompts}.}

\definecolor{close}{rgb}{0.98, 0.93, 0.93} 
\definecolor{open}{rgb}{0.96, 0.99, 0.99} 
\definecolor{rm}{rgb}{1, 1, 0.85}
\definecolor{ft}{rgb}{0.95, 0.85, 0.98}

\definecolor{close}{rgb}{1,1,1} 
\definecolor{open}{rgb}{1,1,1} 
\definecolor{rm}{rgb}{1,1,1}
\definecolor{ft}{rgb}{1,1,1}
\definecolor{closerow}{rgb}{0.3,0.3,1} 
\definecolor{openrow}{rgb}{0.5,0.3,0.1} 
\definecolor{ftrow}{rgb}{0.3,0.5,0.5}
\definecolor{rmrow}{rgb}{1,0,0}

\begin{table}[t!]
\small
\centering
\addtolength{\tabcolsep}{-0.7pt} 
\begin{tabular}{@{}lccccc@{}}
\toprule
\textbf{Model} & \textbf{\natshort} & \textbf{\advshort} & \textbf{\mtshort}  & \textbf{\insshort} & \textbf{\texttt{Avg.}}\\
\midrule
\rowcolor{closerow!25}\multicolumn{6}{l}{Proprietary LLMs}  \\ 
\rowcolor{close} gpt-4o-24-08-06  & \textbf{97.5}    & 84.5                 & 79.8          & \textbf{81.3}   & \textbf{85.7} \\
\rowcolor{close} o1-mini-24-09-12 & 92.5             & \textbf{88.6}        & 79.0          & \textbf{81.3}   & 85.3          \\
\rowcolor{close} gpt-4-0613         & 95.5             & 79.3                 & \underline{81.5}          & 80.4            & 84.2          \\
\rowcolor{close} gpt-4o-24-05-13  & 95.5             & 80.7                 & 79.5          & 80.3            & 84.0          \\
\rowcolor{close} claude-3.5-sonnet  & 91.0             & 81.2                 & 78.5          & 77.5            & 82.0          \\
\rowcolor{close} claude-3-opus      & 94.0             & 76.8                 & 75.5          & 74.1            & 80.1          \\
\rowcolor{close} mistral-large      & 90.0             & 72.1                 & 79.0          & 78.5            & 79.9          \\
\rowcolor{close} gemini-1.5-pro     & 87.0             & 74.9                 & 78.5          & 75.7            & 79.0          \\
\rowcolor{close} gemini-1.5-flash   & 87.5             & 71.3                 & 77.8          & 77.5            & 78.5          \\
\rowcolor{close} gpt-4o-mini        & 88.5             & 68.3                 & 80.2          & 76.6            & 78.4          \\
\rowcolor{close} gemini-1.0-pro     & 85.5             & 54.5                 & 70.8          & 68.7            & 69.9          \\
\rowcolor{close} gpt-3.5-turbo-0125 & 82.5             & 36.4                 & 72.8          & 63.5            & 63.8          \\
\rowcolor{close} claude-3-haiku     & 76.0             & 42.9                 & 68.8          & 62.8            & 62.6          \\ 
\rowcolor{openrow!20}\multicolumn{6}{l}{Open-source LLMs}  \\ 
\rowcolor{open} llama-3.1-405b     & \underline{94.0}             & \underline{83.1}                 & 81.5          & \underline{79.6}            & \underline{84.5}          \\
\rowcolor{open} llama-3.1-70b      & 90.5             & 79.3                 & 82.2 & 79.4            & 82.9          \\
\rowcolor{open} llama-3-70b        & 87.0             & 72.7                 & 80.0          & 78.6            & 79.6          \\
\rowcolor{open} qwen-2-72b         & 92.5             & 69.4                 & 82.2 & 73.1            & 79.3          \\
\rowcolor{open} qwen-2.5-72b       & 90.5             & 67.7                 & \textbf{82.5} & 74.1            & 78.7          \\
\rowcolor{open} qwen-1.5-72b       & 88.5             & 59.7                 & 75.0          & 69.2            & 73.1          \\
\rowcolor{open} glm-4-9b           & 86.0             & 55.0                 & 73.5          & 73.4            & 72.0          \\
\rowcolor{open} yi-1.5-34b         & 86.5             & 56.6                 & 73.8          & 66.9            & 70.9          \\
\rowcolor{open} tulu-2-dpo-70b     & 85.5             & 58.9                 & 73.2          & 66.1            & 70.9          \\
\rowcolor{open} tulu-2-70b         & 86.5             & 58.0                 & 74.5          & 64.7            & 70.9          \\
\rowcolor{open} mixtral-8x7b       & 80.5             & 58.9                 & 73.0          & 68.7            & 70.3          \\
\rowcolor{open} yi-1.5-9b          & 85.0             & 59.1                 & 72.5          & 63.1            & 69.9          \\
\rowcolor{open} qwen-1.5-32b       & 85.5             & 47.3                 & 76.8          & 66.2            & 68.9          \\
\rowcolor{open} llama-3.1-8b       & 78.0             & 50.9                 & 72.5          & 66.5            & 67.0          \\
\rowcolor{open} llama-2-70b        & 80.0             & 32.4                 & 72.2          & 66.9            & 62.9          \\
\rowcolor{open} llama-3-8b         & 70.5             & 43.6                 & 72.5          & 61.7            & 62.1          \\
\rowcolor{open} mistral-7b-v0.3    & 64.5             & 48.0                 & 66.3          & 60.6            & 59.8          \\
\rowcolor{open} tulu-2-dpo-13b     & 67.0             & 38.6                 & 65.5          & 61.2            & 58.1          \\
\rowcolor{open} tulu-2-13b         & 65.5             & 38.6                 & 65.5          & 61.8            & 57.8          \\
\rowcolor{open} llama-2-13b        & 65.0             & 36.4                 & 66.8          & 60.8            & 57.2          \\
\rowcolor{open} tulu-2-dpo-7b      & 56.0             & 43.4                 & 58.5          & 58.9            & 54.2          \\
\rowcolor{open} gemma-7b           & 52.5             & 39.3                 & 64.5          & 57.4            & 53.4          \\
\rowcolor{open} tulu-2-7b          & 45.5             & 46.9                 & 55.2          & 57.8            & 51.4          \\
\rowcolor{open} llama-2-7b         & 42.5             & 49.5                 & 52.0          & 56.4            & 50.1          \\
\rowcolor{open} gemma-2b           & 42.5             & 44.8                 & 54.5          & 56.6            & 49.6          \\
\rowcolor{rmrow!15}\multicolumn{6}{l}{Reward Models}  \\ 
\rowcolor{rm} offsetbias-rm      & 93.0             & 77.1                 & 81.0          & 74.0            & 81.3          \\
\rowcolor{rm} nemotron-4-340b & 95.0             &  84.6        & 75.5          & 69.3            & 81.1          \\
\rowcolor{ftrow!25}\multicolumn{6}{l}{Fine-tuned LLMs}  \\ 
\rowcolor{ft} offsetbias-lm      & 88.0             & 79.9                 & 80.0          & 74.8            & 80.7          \\
\rowcolor{ft} prometheus-2  & 83.0             & 37.3                 & 76.0          & 64.4            & 65.2          \\
\midrule
avg. & 80.7    & 60.2      & 73.4 & 69.3   & 70.9  \\
\bottomrule
\end{tabular}
\addtolength{\tabcolsep}{+0.7pt} 
\caption{Evaluation accuracy of various base LLMs with the base evaluation protocol.
Models are ordered by their performance.
\natshort is \llmbarnatural, \advshort is \adversarial, \mtshort is \mtbench, \insshort is \instrusum. Best column performance is bolded, and best group performance is underlined.
}
\label{tab:baseline} 
\end{table}

Table~\ref{tab:baseline} presents the evaluation accuracy of 38 proprietary and open-source LLMs, together with two state-of-the-art reward models, \texttt{nemotron-4-340b-rm}~\citep{adler2024nemotron} and \texttt{offsetbias-rm}~\citep{park2024offsetbias}, and two strong fine-tuned LLM-evaluators, \texttt{prometheus-2-8x7b}~\cite{kim2024prometheus} and \texttt{offsetbias-lm}~\citep{park2024offsetbias} as baselines.
The model information is in Appendix~\ref{appx_model_details} at Table~\ref{tab:appx_model_registry}.
We note the following observations:

\noindent (1) \textbf{Proprietary vs. Open-Source}: the open-sourced \texttt{llama-3.1-405b} outperforms most of the proprietary LLMs, and \texttt{llama-3.1-70b}  lags slightly behind \texttt{gpt-4o} and \texttt{gpt-4-0613}.

\noindent (2) \textbf{Performance Gap}:
The LLMs at the lower end, such as \texttt{llama-2-7b} and \texttt{gemma-2b}, achieve an accuracy near 50\%, comparable to a random oracle.
On the other hand, \texttt{llama-3.1-405b} achieves a high accuracy of approximately 84\%.

\noindent (3) \textbf{Dataset Difficulty}: 
There is also a large difference in the average LLM performance across different datasets.
For example, the average evaluation accuracy on \llmbarnatural is around 20\% higher than \adversarial.

\noindent (4) \textbf{Comparisons with Reward Models and Fine-tuned LLMs}. 
The strongest LLM-evaluators outperform the state-of-the-art reward models and fine-tuned LLM-evaluators.
The fine-tuned LLM-evaluator, \texttt{offsetbias-lm}, shows a significant improvement over its base model, \texttt{llama-3-8b}, suggesting the potential of fine-tuned LLM-evaluators.
Meanwhile, \texttt{prometheus-2-8x7b} only outperforms its base model (\texttt{mixtral-8x7b}) on the easier datasets \llmbarnatural and \mtbench, indicating a lack of robustness.

These baseline results show that the top open-source LLMs already approach the performance of their proprietary counterparts and offer a wide performance spectrum. 
Therefore, for transparency and reproducibility, we will use mostly open-source LLMs in the rest of our evaluations.

\subsection{Baselines for Evaluation Protocols}
\label{baselines_protocols}

We now establish a baseline for evaluation protocols, which define how the base LLM is used to perform the evaluation.
To this end, we evaluate the evaluation protocols used in three automatic LLM benchmarks for instruction-following -- AlpacaEval~\cite{alpaca_eval}, ArenaHard~\cite{li2024crowdsourced}, and WildBench~\cite{lin2024wildbench}.\footnote{The prompt templates are in Appendix~\ref{appx_benchmark_prompts}.}
Each of these benchmarks uses their evaluation protocol together with a strong base LLM, e.g., GPT-4~\citep{achiam2023gpt}, to perform \textit{pairwise} comparison of different LLMs' outputs.
The individual comparison results are then aggregated to produce a performance ranking of various LLMs.
We note that the efficacy of these benchmarks is evaluated at the \textit{system level}, where their produced ranking is compared against the system ranking from human evaluation benchmarks, e.g., ChatBot Arena~\cite{chiang2024chatbot}.
In contrast, here we aim to evaluate the performance of their evaluation protocols at the \textit{instance level}, measuring their evaluation accuracy against human annotations at individual data instances.

In Table~\ref{tab:benchmark-protocol}, the benchmark evaluation protocols are compared against the base protocol~\cite{zeng2024evaluating} used in \S\ref{subsec:llm-baseline}, where they are used together with the 25 open-source base LLMs evaluated in \S\ref{subsec:llm-baseline} and the strongest proprietary LLM, \texttt{gpt-4o}.
It shows that the benchmark protocols cannot outperform the base protocol, especially on the more challenging  \adversarial and \instrusum datasets.
This indicates that the complex design of the benchmark protocols, which often includes detailed instructions on the evaluation plan and output structure, cannot improve the LLM-evaluators performance at the instance level.
In the next section, we will provide a further examination of various evaluation protocols.

\begin{table}[t!]
\small
\centering
\addtolength{\tabcolsep}{+2pt} 
\begin{tabular}{@{}lccccc@{}}
\toprule
\textbf{Protocol} & \textbf{\natshort} & \textbf{\advshort} & \textbf{\mtshort}  & \textbf{\insshort} & \textbf{\texttt{Avg.}}\\
\midrule
\multicolumn{6}{c}{Average across 25 Open-Source Base LLMs} \\
\midrule
base       & 74.7             & \textbf{53.5}        & 70.7          & \textbf{66.0}   & \textbf{66.2} \\
 arena-hard & \textbf{76.3}    & 46.2                 & \textbf{72.1} & 64.5            & 64.8          \\
 wild-bench  & 74.9             & 47.5                 & 70.7          & 63.0            & 64.0          \\
 alpaca-eval & 65.3             & 49.8                 & 63.1          & 59.4            & 59.4          \\
\midrule
\multicolumn{6}{c}{gpt-4o-2024-0806 as Base LLM} \\
\midrule
base       & \textbf{97.5}    & \textbf{84.5}        & 79.8          & \textbf{81.3}   & \textbf{85.7} \\
 arena-hard & 94.5             & 78.8                 & 83.2          & 75.4            & 83.0          \\
 wild-bench  & 95.5             & 75.5                 & \textbf{84.0} & 73.7            & 82.2          \\
 alpaca-eval & 94.0             & 70.2                 & 83.7          & 76.5            & 81.1          \\
\bottomrule
\end{tabular}
\addtolength{\tabcolsep}{-2pt} 
\caption{Evaluation accuracy of evaluation protocols in existing LLM benchmarks compared against the base evaluation protocol.  
}
\label{tab:benchmark-protocol} 
\end{table}

\section{Evaluating Evaluation Protocols}
\label{sec:protocols}

In \S\ref{sec:status-quo}, we only tested the LLM-evaluators with the base and benchmark evaluation protocols. We now expand the evaluation dimensions to include various protocols proposed in recent work. By using 25 open-source LLMs, we believe this evaluation will provide a fairer and more rigorous examination.

\subsection{Evaluation Protocols}
\label{sec:all_protocols}
In our evaluation, we examine 15 protocols derived from previous work. 
To address the unavailability of some prompt templates and to ensure a fair comparison, we design prompt templates ourselves when necessary.
We ensure that all prompt templates adhere to unified formatting and style, and we refine them iteratively to make sure that the protocols can perform to their full potential.
The evaluated protocols are outlined below, with their prompt templates provided in Appendix~\ref{appx_prompts}.

\paragraph{Baseline Protocol}
\noindent (1) \texttt{base}: the vanilla approach used in \S\ref{sec:status-quo} which directly predicts the pairwise comparison outcome, proposed in \citet{zeng2024evaluating}. 

\paragraph{Enhanced Protocols} Five other protocols from \citet{zeng2024evaluating} are evaluated, which include various enhancements based on the \texttt{base} protocol:

\noindent (2) \texttt{cot}: the LLM is asked to provide a chain-of-thought~\cite{wei2022chain} explanation before making the final decision.

\noindent (3) \texttt{metric}: the LLM is prompted to generate a few metrics for the evaluation task first, which are later used in the actual evaluation.

\noindent (4) \texttt{reference}: the LLM is prompted to generate a ``reference'' output for the given instruction, which is later used in the actual evaluation.

\noindent (5) \texttt{metric+reference}: a combination of the \texttt{metric} and \texttt{reference} methods.

\noindent (6) \texttt{swap\&synthesize}: based on \texttt{cot} and inspired by \citet{du2024improving}, this method requires the LLM to resolve self-disagreement in predictions from two output orders and make a final decision.

\paragraph{Complex Protocols}
Beyond the enhanced protocols, 7 complex protocols are evaluated based on evaluation methods proposed in previous work.

\noindent (7) \texttt{fine-grained-diff}: Similar to~\citet{min-etal-2023-factscore}, this protocol guides the LLM to first identify \textit{fine-grained differences} in output pairs and then provide a detailed rationale for choosing the better output considering these differences.

\noindent (8) \texttt{multi-role-round1} \& (9) \texttt{multi-role-\\round2}: 
Inspired by frameworks that use multiple agents as judges~\citep{li2023prd, chan2024chateval, zhao2024autoarena}, these two protocols use a \textit{multi-role debate} setting where multiple evaluators are instantiated from an LLM using prompts with specific role descriptions, to bring diverse perspectives to the evaluation process.
The evaluators will generate their responses sequentially, potentially in \textit{multiple rounds}, to engage in a debate that leads to the final prediction.  
We evaluate its single-round and two-round variants. 

\noindent (10) \texttt{multi-aspect-two} \& (11) \texttt{multi-aspect-\\single}: Similar to several related studies~\citep{saha2023branchsolvemerge, gong2023coascore, li2023generative, li2024decompose}, this protocol performs a \textit{multi-aspect comparison} of the output pairs, with two variants:
the \textit{two-stage} protocol prompts the LLMs to evaluate each quality aspect in separate inference passes, while the \textit{single-stage} protocol requires the LLMs to conduct a multi-aspect evaluation in a single inference pass before making the final prediction.

\noindent (12) \texttt{gpt4-reference}: Similar to the \texttt{reference} protocol, this protocol uses a reference output generated by \texttt{gpt-4o} to assist the evaluation.

\noindent (13) \texttt{prepair}: Adapted from \citet{jeong2024prepair}, this protocol incorporates pointwise reasoning within a pairwise evaluation framework, leveraging the robustness of pointwise evaluation against biases while maintaining the comparative benefits of pairwise evaluation. 

\paragraph{Self-consistency Protocols}
Self-consistency is a commonly used decoding approach that can improve the LLMs' performance in various reasoning tasks~\cite{wang2023selfconsistency}.
Used together with CoT prompting, self-consistency generates the final prediction by taking a majority vote on the predictions made in each generation pass.

\noindent (14) \texttt{cot\&self-consistency}:
Self-consistency in pairwise comparison determines the more frequently preferred output. 
We use a sampling temperature of 0.7 and generate 9 CoTs for voting.

\noindent (15) \texttt{protocol-consistency}:
Beyond different CoTs, a majority vote can be applied across various evaluation protocols.
We evaluate this approach using the 5 \textit{enhanced} protocols.

\subsection{Results}
\label{subsec:protocol-overall}

\begin{table}[t!]
\small
\centering
\addtolength{\tabcolsep}{-4.0pt} 
\begin{tabular}{lrrrrr}
\toprule
\textbf{Protocol} & \textbf{\natshort} & \textbf{\advshort} & \textbf{\mtshort}  & \textbf{\insshort} & \textbf{\texttt{Avg.}}\\
\midrule
  prepair (13)              & \bettersig76.4          & \bettersig\textbf{61.8} & 69.7                    & \worsesig63.8   & \bettersig\textbf{67.9} \\
 gpt4-reference (12)       & \bettersig76.7          & \bettersig58.0          & 70.1                    & 66.0            & \bettersig67.7          \\
 metric+reference (5)      & \bettersig76.6          & \bettersig58.3          & \worsesig70.0           & 65.6            & \bettersig67.6          \\
 protocol-consistency (15) & \bettersig76.3          & \bettersig55.9          & 70.9                    & 66.1            & \bettersig67.3          \\
 metric (3)                & \bettersig75.8          & \bettersig56.2          & 70.7                    & 65.7            & \bettersig67.1          \\
 reference (4)             & \bettersig76.2          & \bettersig57.5          & \worsesig69.4           & \worsesig65.2   & \bettersig67.1          \\
 swap\&synthesize (6)      & 75.6                    & \bettersig54.4          & 70.8                    & \textbf{66.2}   & \bettersig66.8          \\
 cot\&consistency (14)     & 74.9                    & 54.1                    & 70.5                    & \worsesig65.4   & 66.2                    \\
 base (1)                  & 74.7                    & 53.5                    & 70.7                    & 66.0            & 66.2                    \\
 cot (2)                   & \worsesig73.6           & 53.6                    & 70.2                    & \worsesig64.9   & \worsesig65.6           \\
 multi-aspect-two (10)     & \bettersig\textbf{77.1} & \worsesig42.3           & \bettersig\textbf{72.5} & \worsesig62.3   & \worsesig63.6           \\
 fine-grained-diff (7)     & \worsesig71.2           & \worsesig49.5           & \worsesig69.2           & \worsesig61.8   & \worsesig62.9           \\
 multi-role-round1 (8)     & \worsesig68.0           & 53.9                    & \worsesig66.2           & \worsesig61.6   & \worsesig62.4           \\
 multi-role-round2 (9)     & \worsesig68.4           & 53.4                    & \worsesig65.7           & \worsesig61.7   & \worsesig62.3           \\
 multi-aspect-single (11)  & \worsesig69.6           & \worsesig40.8           & 70.5                    & \worsesig62.4   & \worsesig60.8           \\
\bottomrule
\end{tabular}
\addtolength{\tabcolsep}{+4.0pt} 
\caption{Average evaluation accuracy of different evaluation protocols across various LLMs, ordered by their average performance.
The protocol indexes introduced in \S\ref{sec:all_protocols} are in parentheses.
\bettersig, \worsesig: significantly better or worse than the base protocol ($p<0.05$).
}
\label{tab:protocol-avg} 
\end{table}

\paragraph{Evaluation Accuracy} Table~\ref{tab:protocol-avg} demonstrates the evaluation accuracy of various protocols averaged over different base LLMs.
We note the following:

\noindent (1) \texttt{prepair} and \texttt{gpt4-reference} achieve the strongest average performance, achieving a 1.7\% higher accuracy compared to the base protocol.

\noindent (2) \texttt{multi-aspect-two} achieves the best performance on \llmbarnatural and \mtbench.
However, its performance on \adversarial ranks among the worst.
This highlights that the protocol performance can significantly vary across different datasets.

\noindent (3) On average, most complex protocols fail to outperform the base protocol, despite their higher computational costs and multi-step, fine-grained nature, indicating that designing a robust, superior evaluation protocol is a non-trivial task.

\noindent (4) Similarly, the approaches that have been proven effective on various reasoning tasks, chain-of-thought (\texttt{cot}) and self-consistency (\texttt{cot\&consistency}), also fail to bring significant improvement over the \texttt{base} protocol.

\renewcommand{\arraystretch}{1.12} 
\begin{table}[t!]
\small
\centering
\addtolength{\tabcolsep}{-2.5pt} 
\begin{tabular}{@{}lccccc@{}}
\toprule
\textbf{Protocol} & \textbf{\natshort} & \textbf{\advshort} & \textbf{\mtshort}  & \textbf{\insshort} & \textbf{\texttt{Avg.}}\\
\midrule
 swap\&synthesize (6)      & \textbf{89.3}    & \textbf{84.4}        & \textbf{86.0} & \textbf{86.1}   & \textbf{86.5} \\
 prepair (13)              & 62.1             & 59.7                 & 59.3          & 38.9            & 55.0          \\
 multi-aspect-two (10)     & 62.4             & 51.9                 & 54.6          & 36.6            & 51.4          \\
 fine-grained-diff (7)     & 53.1             & 45.3                 & 54.6          & 33.5            & 46.6          \\
 protocol-consistency (15) & 56.5             & 47.0                 & 50.9          & 30.7            & 46.3          \\
 metric (3)                & 53.6             & 48.4                 & 52.7          & 29.4            & 46.0          \\
 metric+reference (5)      & 57.4             & 48.7                 & 49.7          & 28.3            & 46.0          \\
 base (1)                  & 54.5             & 47.3                 & 49.6          & 32.4            & 46.0          \\
 cot\&consistency (14)     & 54.0             & 42.0                 & 48.1          & 36.9            & 45.2          \\
 multi-aspect-single (11)  & 45.6             & 44.6                 & 50.7          & 35.9            & 44.2          \\
 gpt4-reference (12)       & 51.3             & 47.2                 & 45.9          & 25.7            & 42.5          \\
 cot (2)                   & 47.7             & 38.7                 & 45.7          & 34.6            & 41.7          \\
 reference (4)             & 50.8             & 44.5                 & 44.2          & 24.9            & 41.1          \\
 multi-role-round1 (8)     & 38.3             & 26.8                 & 31.8          & 25.4            & 30.6          \\
 multi-role-round2 (9)     & 37.1             & 25.9                 & 31.8          & 25.9            & 30.2          \\
\bottomrule
\end{tabular}
\addtolength{\tabcolsep}{+2.5pt} 
\caption{Average self-agreement rate of various evaluation protocols across various LLMs (ordered).
The protocol indexes introduced in \S\ref{sec:all_protocols} are in parentheses.
}
\label{tab:protocol-avg-agreement} 
\end{table}
\renewcommand{\arraystretch}{1.1} 

\paragraph{Self-Agreement Rate} Table~\ref{tab:protocol-avg-agreement} shows the protocols self-agreement rates, demonstrating that
(1) \texttt{swap\&synthesize} can significantly enhance the self-agreement rate;
(2) The multi-role debate protocols, \texttt{multi-role-round1\&2}, yield a significantly lower self-agreement rate, indicating that introducing more complex evaluation processes can lead to larger self-inconsistency.

\paragraph{Best LLM-Evaluators}

Table~\ref{tab:best-evaluator} displays the LLM-evaluators that achieve the highest evaluation accuracy on each dataset, together with the evaluation accuracy of the same base LLM achieved with the \texttt{base} protocol.
It shows that \texttt{llama-3.1-405b}, which achieves the strongest performance among the open-source LLMs, remains the strongest base LLM across three datasets.
On the other hand, the evaluation protocols used by the best LLM-evaluators on the four datasets differ, indicating greater variance in their capabilities. 

\begin{table}[t!]
\small
\centering
\addtolength{\tabcolsep}{-4.0pt} 
\begin{tabular}{lllrr}
\toprule
\textbf{Dataset} & \textbf{LLM} & \textbf{Protocol} & \textbf{Acc.}  & \textbf{B.Acc.}\\
\midrule
 \natshort & llama-3.1-405b & \texttt{swap\&synthesize} & 98.0 & 94.0 \\
 \advshort & llama-3.1-405b  & \texttt{gpt4-reference} & 87.8 & 83.1 \\
 \mtshort & qwen-2.5-72b & \texttt{metric+reference} & 84.0  & 82.5 \\
 \insshort & llama-3.1-405b & \texttt{prepair} &  82.7 & 79.6 \\
\bottomrule
\end{tabular}
\addtolength{\tabcolsep}{+4.0pt} 
\caption{Best LLM-evaluators identified on each dataset.
\textbf{Acc.} is the evaluation accuracy of the best LLM-evaluator, \textbf{B.Acc.} is the evaluation accuracy achieved by the same base LLM with the \texttt{base} protocol. 
}
\label{tab:best-evaluator} 
\end{table}

\section{Analysis}
\label{sec:analysis}

In \S\ref{sec:protocols}, a total number of 375 LLM-evaluators are evaluated, combining 25 base LLMs and 15 evaluation protocols. 
We now present detailed analyses of the base LLMs, evaluation protocols, and datasets, using these comprehensive evaluation results to address a series of specific research questions.

\begin{table}[t!]
\small
\centering
\addtolength{\tabcolsep}{+1pt} 
\begin{tabular}{@{}lccccc@{}}
\toprule
\textbf{Model} & \textbf{\natshort} & \textbf{\advshort} & \textbf{\mtshort}  & \textbf{\insshort} & \textbf{\texttt{Avg.}}\\
\midrule
 llama-3.1-405b  & \textbf{94.1}    & \textbf{81.3}        & 81.4          & \textbf{80.0}   & \textbf{84.2} \\
 llama-3.1-70b   & 91.7             & 80.2                 & 81.0          & 75.3            & 82.1          \\
 qwen-2-72b      & 91.6             & 69.9                 & \textbf{82.2} & 72.8            & 79.1          \\
 qwen-2.5-72b    & 89.6             & 71.0                 & 81.2          & 72.4            & 78.6          \\
 llama-3-70b     & 88.2             & 71.5                 & 80.0          & 74.3            & 78.5          \\
 qwen-1.5-72b    & 86.1             & 56.0                 & 76.4          & 68.0            & 71.6          \\
 yi-1.5-34b      & 86.5             & 57.6                 & 73.9          & 66.0            & 71.0          \\
 tulu-2-dpo-70b  & 84.2             & 56.1                 & 73.7          & 66.5            & 70.1          \\
 mixtral-8x7b    & 82.2             & 54.7                 & 73.8          & 66.4            & 69.3          \\
 tulu-2-70b      & 83.6             & 55.0                 & 72.8          & 65.4            & 69.2          \\
 qwen-1.5-32b    & 85.7             & 47.9                 & 77.0          & 64.3            & 68.7          \\
 glm-4-9b        & 79.1             & 54.4                 & 70.6          & 68.2            & 68.1          \\
 yi-1.5-9b       & 77.1             & 55.7                 & 71.3          & 60.1            & 66.1          \\
 llama-3.1-8b    & 75.3             & 54.7                 & 70.3          & 62.3            & 65.7          \\
 llama-3-8b      & 71.8             & 47.9                 & 71.4          & 62.6            & 63.4          \\
 llama-2-70b     & 74.4             & 36.6                 & 68.5          & 63.3            & 60.7          \\
 mistral-7b-v0.3 & 65.1             & 47.3                 & 66.6          & 61.9            & 60.2          \\
 tulu-2-dpo-13b  & 66.0             & 40.0                 & 65.3          & 60.3            & 57.9          \\
 tulu-2-13b      & 63.3             & 39.7                 & 65.4          & 60.2            & 57.2          \\
 llama-2-13b     & 63.9             & 36.3                 & 62.7          & 56.9            & 54.9          \\
 tulu-2-dpo-7b   & 58.2             & 43.7                 & 57.2          & 57.2            & 54.1          \\
 gemma-7b        & 51.9             & 42.7                 & 59.2          & 56.7            & 52.6          \\
 tulu-2-7b       & 49.1             & 45.8                 & 54.9          & 56.9            & 51.7          \\
 llama-2-7b      & 48.6             & 45.3                 & 54.7          & 54.4            & 50.8          \\
 gemma-2b        & 44.6             & 47.0                 & 53.8          & 55.5            & 50.3          \\
\bottomrule
\end{tabular}
\addtolength{\tabcolsep}{-1pt} 
\caption{Evaluation accuracy of different base LLMs averaged over 15 evaluation protocols. The base LLMs are ordered by their average performance.
}
\label{tab:models} 
\end{table}

\subsection{Analysis of Base LLMs}
\label{subsec:llm-analysis}

\paragraph{What is the average performance of the base LLMs across different protocols?}
Table~\ref{tab:models} displays the LLMs' average evaluation accuracy over 15 protocols, showing that \texttt{llama-3.1-70b} is the strongest evaluator on average, while \texttt{qwen-2-72b} achieves the best performance on \mtbench.

\begin{figure}[t!]
    \centering
    \includegraphics[width=1\linewidth]{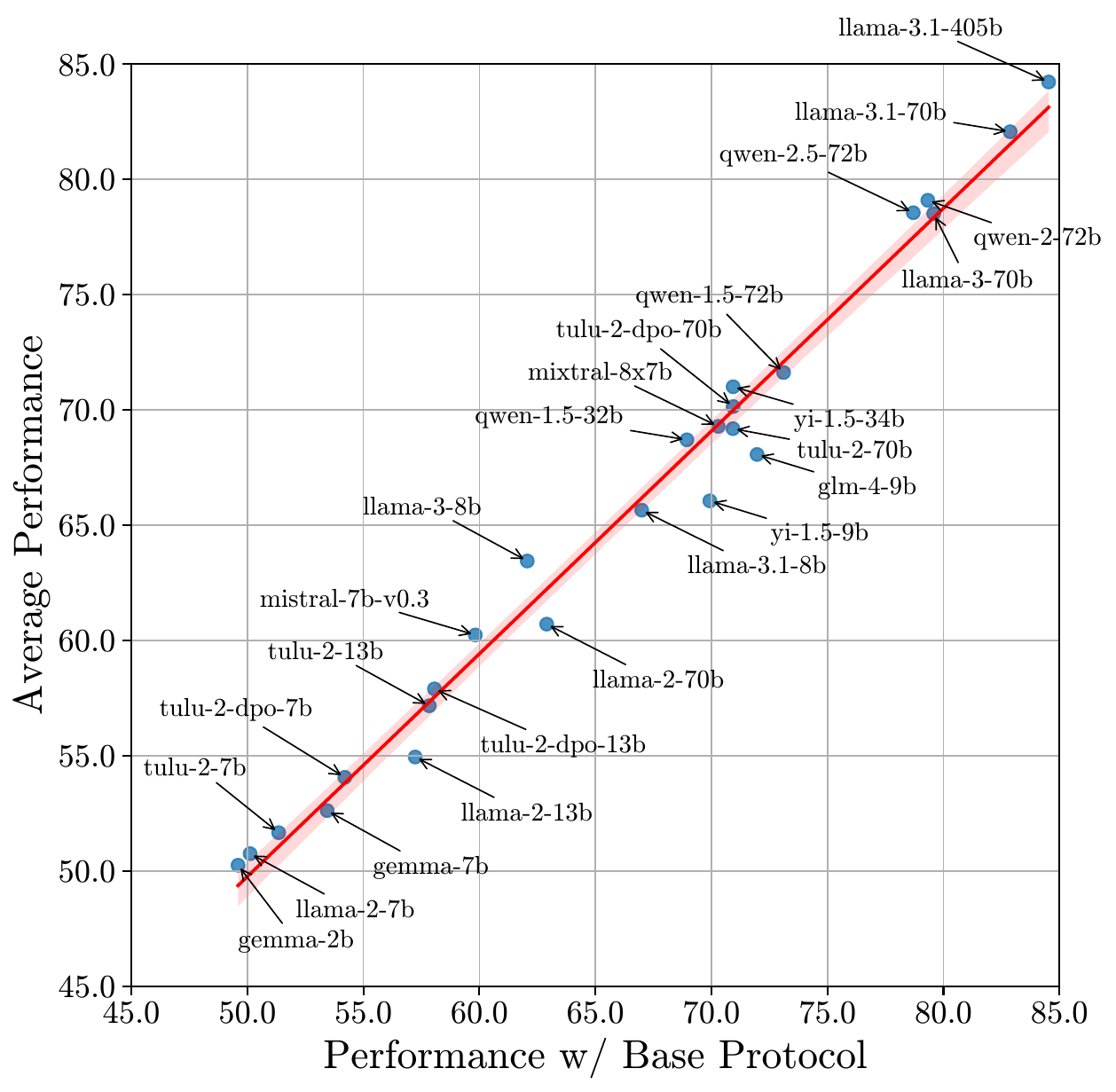}
 \caption{\label{fig:protocol-performance-compare}Correlation between the base LLMs' evaluation accuracy with the \texttt{base} protocol and their average accuracy across 15 protocols. The fitted regression line and the 95\% confidence interval are displayed.
    }
\end{figure}

\paragraph{How does base LLMs' ranking change with different protocols?}
Figure~\ref{fig:protocol-performance-compare} shows the evaluation accuracy of base LLMs achieved with the \texttt{base} protocol and the average accuracy across different protocols.
The results demonstrate a high positive correlation between them, achieving a Spearman's coefficient of 0.983.
This indicates that \textbf{using a single \texttt{base} protocol for the evaluation of the base LLMs' is likely to yield reliable results}.

\begin{table}[t!]
\small
\centering
\addtolength{\tabcolsep}{1pt} 
\begin{tabular}{@{}lrrrrr@{}}
\toprule
\textbf{Model} & \textbf{\natshort} & \textbf{\advshort} & \textbf{\mtshort}  & \textbf{\insshort} & \textbf{\texttt{Avg.}}\\
\midrule
 mistral-7b-v0.3 & 16.0             & 7.8                  & 8.2           & \textbf{5.5}    & \textbf{9.4} \\
 llama-2-7b      & \textbf{23.5}    & 0.2                  & \textbf{11.7} & 1.2             & 9.2          \\
 tulu-2-7b       & 21.0             & 2.7                  & 10.3          & 1.5             & 8.8          \\
 llama-3-8b      & 9.5              & 17.9                 & 3.0           & 4.5             & 8.7          \\
 yi-1.5-34b      & 5.0              & 17.6                 & 4.5           & 3.6             & 7.7          \\
 tulu-2-dpo-13b  & 10.0             & 11.8                 & 6.0           & 1.7             & 7.4          \\
 llama-3.1-8b    & 5.0              & \textbf{18.5}        & 3.3           & -0.7            & 6.5          \\
 tulu-2-dpo-7b   & 15.5             & 5.2                  & 5.2           & 0.0             & 6.5          \\
 qwen-1.5-32b    & 4.0              & 16.5                 & 2.0           & 3.3             & 6.4          \\
 tulu-2-dpo-70b  & 4.0              & 11.0                 & 3.0           & 3.5             & 5.4          \\
 tulu-2-13b      & 5.0              & 10.7                 & 4.5           & 1.0             & 5.3          \\
 glm-4-9b        & 1.0              & 14.1                 & 5.0           & 0.9             & 5.2          \\
 qwen-2.5-72b    & 2.5              & 13.6                 & 1.5           & 3.2             & 5.2          \\
 mixtral-8x7b    & 6.0              & 7.4                  & 4.2           & 1.1             & 4.7          \\
 gemma-7b        & 8.0              & 9.7                  & -0.2          & 0.6             & 4.5          \\
 llama-2-70b     & 2.5              & 11.8                 & 0.5           & 3.3             & 4.5          \\
 tulu-2-70b      & 4.0              & 8.5                  & 0.3           & 5.2             & 4.5          \\
 llama-3-70b     & 4.5              & 8.9                  & 2.2           & 2.1             & 4.4          \\
 qwen-1.5-72b    & 1.0              & 8.6                  & 4.0           & 0.7             & 3.6          \\
 gemma-2b        & 6.5              & 5.6                  & 2.0           & 0.1             & 3.6          \\
 qwen-2-72b      & 2.0              & 8.8                  & 1.2           & 1.8             & 3.5          \\
 llama-2-13b     & 6.0              & 6.0                  & 1.5           & 0.2             & 3.4          \\
 llama-3.1-405b  & 4.0              & 4.7                  & 0.8           & 3.2             & 3.2          \\
 llama-3.1-70b   & 4.5              & 5.5                  & 0.2           & 1.2             & 2.9          \\
 yi-1.5-9b       & 0.5              & 5.6                  & 2.3           & 1.6             & 2.5          \\
\bottomrule
\end{tabular}
\addtolength{\tabcolsep}{-1pt} 
\caption{Optimal evaluation accuracy improvement (Eq.~\ref{eq:improvement}) of base LLMs achieved by the most compatible evaluation protocols, ordered by average improvement.
}
\label{tab:models-improvement} 
\end{table}

\paragraph{How large is the optimal improvement gained from different evaluation protocols for base LLMs?}

Table~\ref{tab:models-improvement} displays the optimal evaluation accuracy improvement ($\Tilde{s}$) achieved by different evaluation protocols over the \texttt{base} protocol for various base LLMs.
That is,
\begin{equation}
\label{eq:improvement}
    \Tilde{s} = \max_{p \in \mathcal{P}} s(p) - s(\hat{p}),
\end{equation}
where $s(p)$ is the evaluation accuracy of an evaluation protocol $p$, $\hat{p}$ denotes the \texttt{base} protocol, $\mathcal{P}$ is the set of protocols excluding $\hat{p}$.
The results indicate that \textbf{less capable LLMs are more likely to achieve larger improvements when the suitable protocols are used}, showing a -0.455 Spearman's correlation between the base LLMs' performance with the \texttt{base} protocol and their optimal performance with the most compatible evaluation protocol.
We hypothesize this is because the inductive biases and constraints introduced by the more complicated protocols help less capable LLMs overcome their potential biases and limitations. 

\subsection{Analysis of Evaluation Protocols}
\label{subsec:analysis-protocols}

\begin{table}[t!]
\small
\centering
\addtolength{\tabcolsep}{-2pt} 
\begin{tabular}{@{}lccccc@{}}
\toprule
\textbf{Protocol} & \textbf{\natshort} & \textbf{\advshort} & \textbf{\mtshort}  & \textbf{\insshort} & \textbf{\texttt{Avg.}}\\
\midrule
 metric+reference (3)    & 95.0             & 86.2                 & \textbf{84.0} & 82.4            & \textbf{86.9} \\
 reference  (6)          & 97.5             & 85.9                 & 83.2          & 80.9            & \textbf{86.9} \\
 swap\&synthesize (7)    & \textbf{98.0}    & 84.8                 & 82.7          & 80.4            & 86.5          \\
 gpt4-reference (2)      & 94.5             & \textbf{87.8}        & 82.5          & 81.0            & 86.4          \\
 cot\&consistency (8)    & 96.5             & 84.2                 & 82.2          & 81.5            & 86.1          \\
 protocol-consistency (4) & 95.0             & 85.0                 & 82.8          & 81.4            & 86.0          \\
 prepair (1)             & 94.5             & 84.8                 & 81.2          & \textbf{82.7}   & 85.8          \\
 metric (5)              & 95.0             & 82.1                 & 83.5          & 80.9            & 85.4          \\
 cot (10)                 & 96.0             & 82.8                 & 82.2          & 79.7            & 85.2          \\
 base  (9)               & 94.0             & 83.1                 & 82.5          & 79.6            & 84.8          \\
 multi-role-round2 (14)   & 94.0             & 81.5                 & 81.5          & 79.3            & 84.1          \\
 fine-grained-diff (12)    & 92.5             & 77.9                 & 83.0          & 79.4            & 83.2          \\
 multi-role-round1 (13)   & 92.0             & 81.7                 & 80.2          & 76.8            & 82.7          \\
 multi-aspect-two  (11)   & 93.0             & 73.0                 & 83.0          & 79.0            & 82.0          \\
 multi-aspect-single (15)  & 87.5             & 65.8                 & 83.0          & 75.8            & 78.0          \\ 
\bottomrule
\end{tabular}
\addtolength{\tabcolsep}{+2pt} 
\caption{Optimal evaluation accuracy of different evaluation protocols with the most compatible base LLMs, ordered by their performance.
The protocols' average performance rankings (Table~\ref{tab:protocol-avg}) are in parentheses.
}
\label{tab:protocol-best} 
\end{table}

\paragraph{What is the evaluation protocols' optimal performance?}
In Table~\ref{tab:protocol-avg} of \S\ref{subsec:protocol-overall}, the evaluation protocols' performance is evaluated across all base LLMs.
Table~\ref{tab:protocol-best} instead shows the optimal performance of evaluation protocols, i.e., their evaluation accuracy with the most compatible base LLM.
The results show that the \textbf{evaluation protocols' optimal performance can significantly differ from their average performance}.
For example, while \texttt{prepair} achieves the best average performance, it ranks only 7th in terms of optimal performance.
Moreover, the Spearman's correlation between the rankings of average and optimal performance for the evaluation protocols is 0.789, much lower than the correlation between rankings of average and optimal performance for the base LLMs (0.977).

\paragraph{How do base LLMs' capabilities affect evaluation protocol's performance?}

\begin{figure}[t!]
    \centering
    \includegraphics[width=1\linewidth]{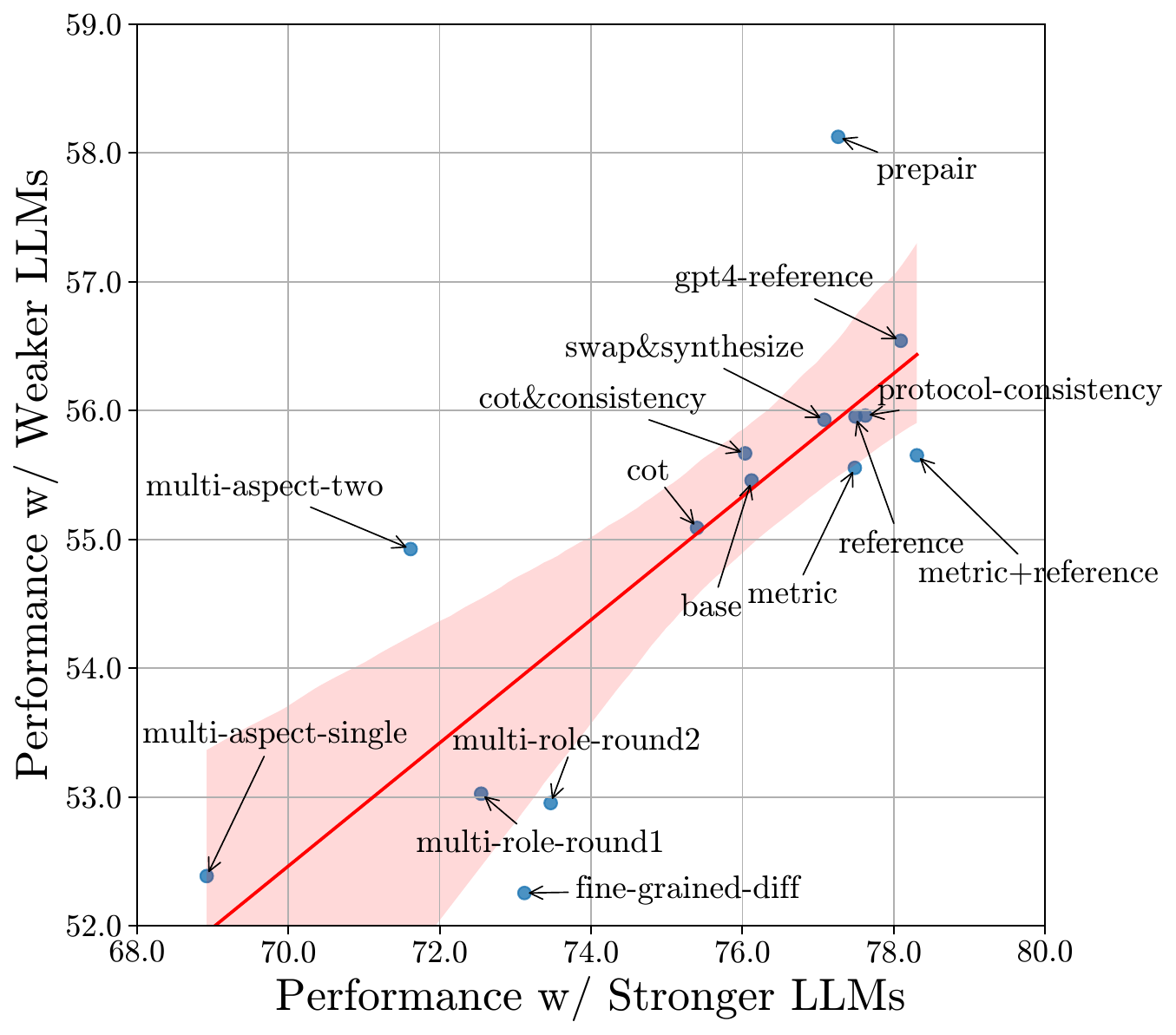}
 \caption{\label{fig:protocol-performance-group}Evaluation protocols' evaluation accuracy with the stronger and weaker base LLM groups, with a fitted regression line and a 95\% confidence interval.
    }
\end{figure}

The previous analysis shows that the evaluation protocol's performance can be significantly affected by the base LLMs used.
Therefore, we provide a further examination of the protocol performance with two groups of LLMs: one containing the strongest 10 LLMs identified in Table~\ref{tab:models}, and another containing the weakest 10.
Figure~\ref{fig:protocol-performance-group} demonstrates a 0.807 Spearman's correlation between the protocol performance ranking with these two groups of LLMs.
We note the effectiveness of evaluation protocols can substantially vary with the capabilities of the base LLMs used.
For example, \texttt{prepair} is significantly better than other protocols for weaker LLMs, while \texttt{metric+reference} works better with stronger LLMs.
This suggests that \textbf{a robust evaluation of evaluation protocols requires multiple base LLMs with a diverse performance range}.

\begin{figure*}[t!]
    \centering
    \begin{subfigure}[b]{0.24\linewidth}
        \centering
        \includegraphics[width=\linewidth]{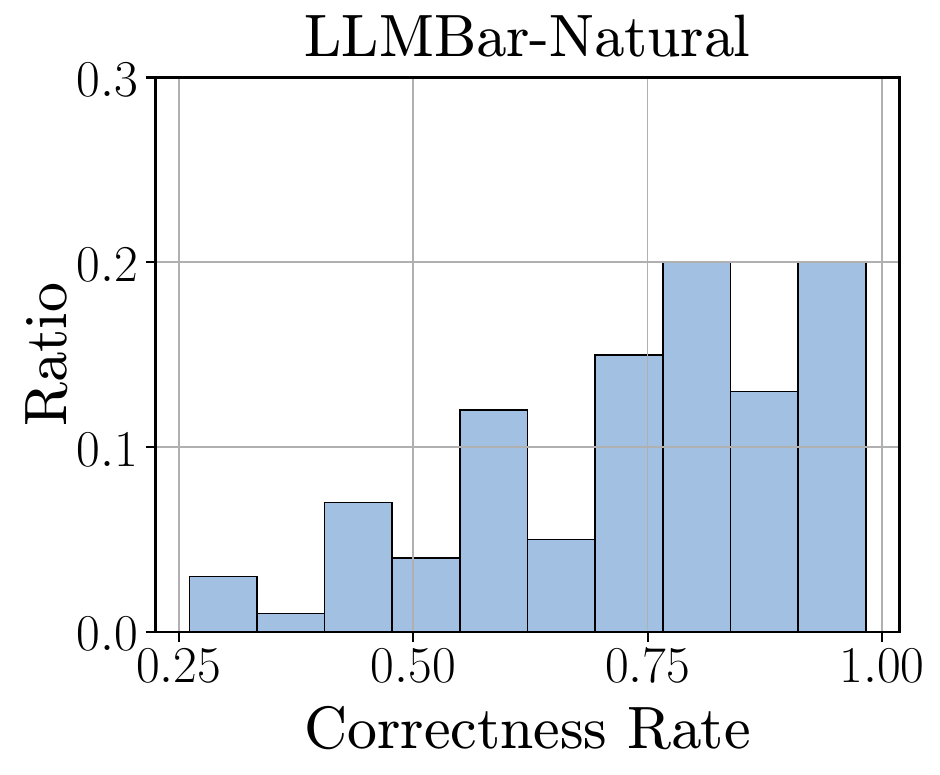}
    \end{subfigure}
    \hfill
    \begin{subfigure}[b]{0.24\linewidth}
        \centering
        \includegraphics[width=\linewidth]{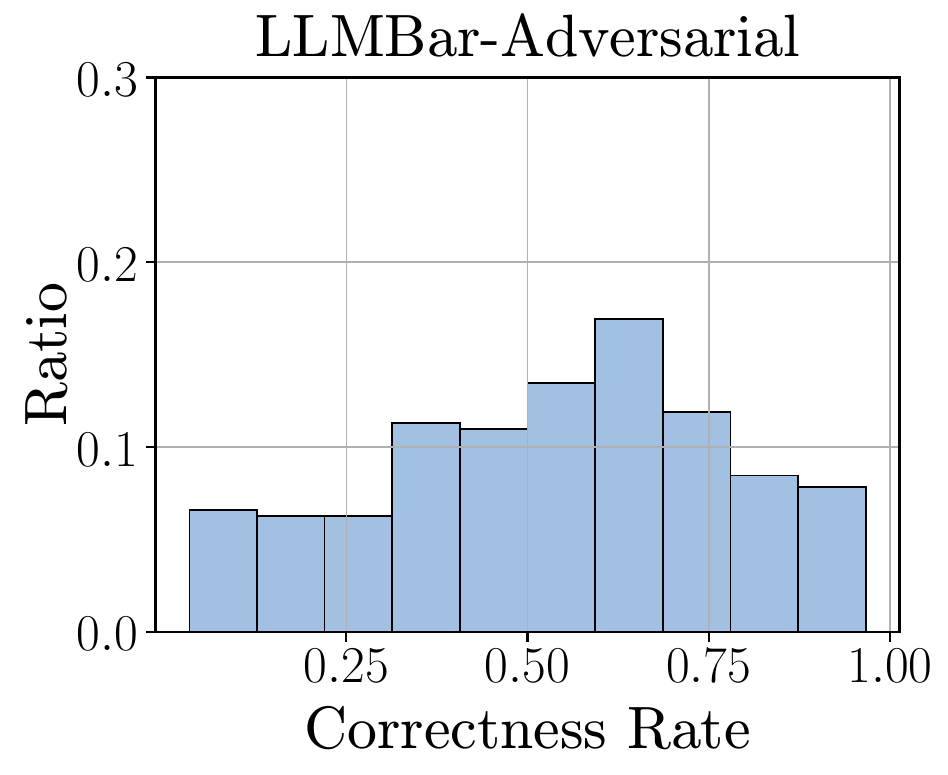}
    \end{subfigure}
    \begin{subfigure}[b]{0.24\linewidth}
        \centering
        \includegraphics[width=\linewidth]{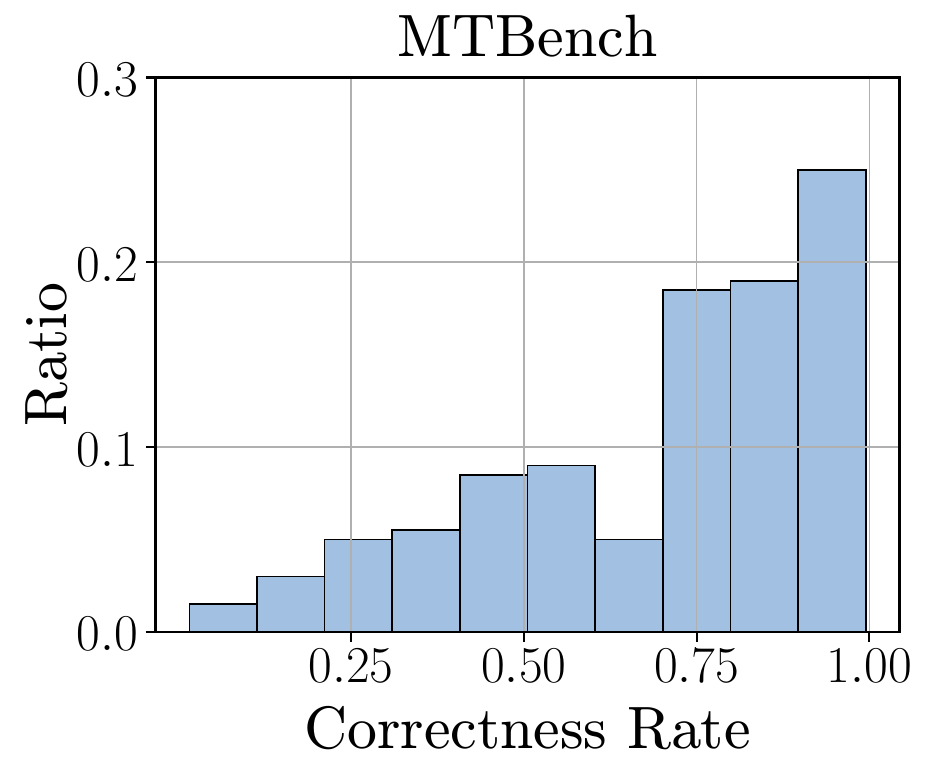}
    \end{subfigure}
    \hfill
    \begin{subfigure}[b]{0.24\linewidth}
        \centering
        \includegraphics[width=\linewidth]{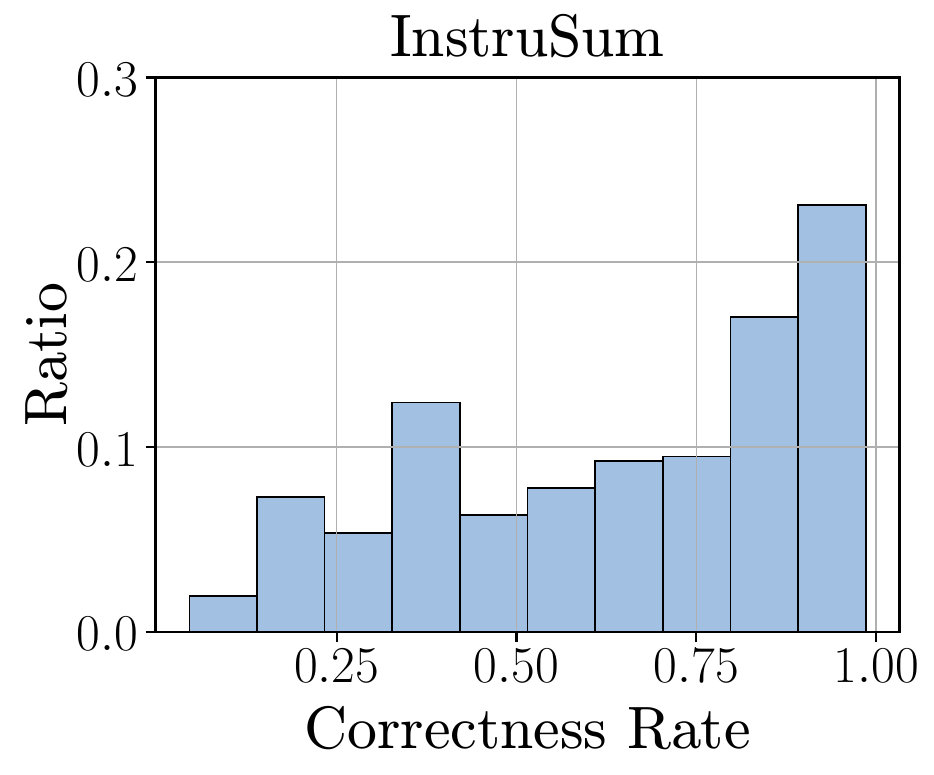}
    \end{subfigure}
    \hfill
    \vspace{-1mm}
    \caption{Distribution of the correctness rate (Eq.~\ref{eq:correctness}) of data examples in each dataset over different LLM-evaluators.}
    \label{fig:correctness-rate}
\end{figure*}

\subsection{Analysis of datasets}
\label{subsec:analysis-dataset}

\paragraph{What is the difficulty level of different datasets?}

The large number of LLM-evaluators in our evaluation allows us to measure the difficulty of individual data examples. Therefore, we calculate the average \textit{correctness rate} of different LLM-evaluators on each example in different datasets, which is defined as the average evaluation accuracy:
\begin{equation}
\label{eq:correctness}
\resizebox{0.5\hsize}{!}{$
 \mathcal{C}(x) = \sum_{i}\frac{\mathrm{Acc}(x;e_i)}{N},
$}
\end{equation}
where $\mathcal{C}(x)$ is the correctness rate of data example $x$, $\mathrm{Acc}(x;e_i)$ is the evaluation accuracy of LLM-evaluator $e_i$, and $N$ is the number of evaluators.
 
Figure~\ref{fig:correctness-rate} shows the distribution of this correctness rate of data examples.
We note:

\noindent (1) \llmbarnatural, \mtbench, and \instrusum have a similar data example difficulty distribution, with a small portion of examples where the LLM-evaluators are rarely correct.

\noindent (2) \adversarial exhibits a different pattern: the distribution peaks at examples of medium difficulty, while the easiest and most difficult examples are similarly proportioned. 
This suggests that different LLM-evaluators may have distinct sets of adversarial examples.

\begin{figure}[t!]
    \centering
    \includegraphics[width=0.6\linewidth]{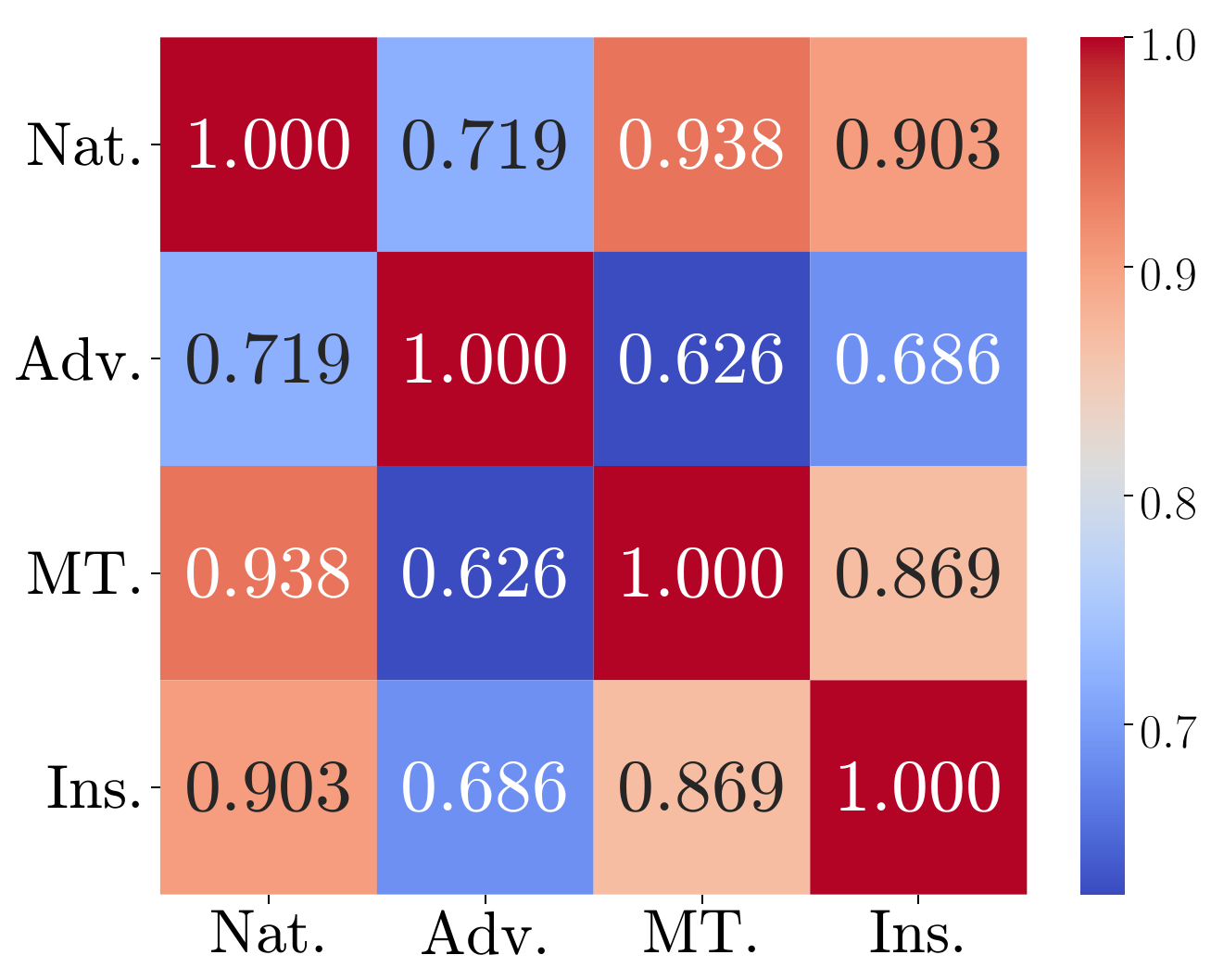}
 \caption{\label{fig:dataset-corr} Spearman's correlations between the performance ranking of LLM-evaluators on different datasets.
    }
\end{figure}

\paragraph{Are LLM-evaluators' rankings consistent over different datasets?}

To better understand how the LLM-evaluators' performance differs across different datasets, in Figure~\ref{fig:dataset-corr} we present the Spearman's correlations between LLM-evaluators' performance ranking between different datasets.
The results show that \llmbarnatural and \mtbench exhibit the highest level of similarity.
In contrast, \adversarial displays a much lower correlation with the other datasets, suggesting the necessity of using multiple datasets for evaluation.

\section{Conclusion}

In this work, we conducted a large-scale meta-evaluation of instruction following, examining 25 open-source base LLMs and 15 evaluation protocols while identifying the best-performing LLM-evaluators over 4 datasets.
We found that a reliable evaluation of base LLMs' evaluation capabilities can likely be achieved with a single evaluation protocol due to the stability of their performance across different protocols.
However, evaluating evaluation protocols should involve a diverse group of base LLMs, as they can significantly impact the evaluation protocols' effectiveness.
We hope that our findings and meta-evaluation suite, \ours, can pave the way for further studies in this direction.

\section*{Limitations}

\noindent \textbf{Evaluation Scope:} 
Our evaluation centered around generic LLMs and evaluation protocols.
As discussed in \S\ref{sec:related-work}, we did not focus on reward models trained to evaluate output quality or LLMs fine-tuned for instruction-following. 
We note that a future study incorporating these systems could yield more comprehensive results.

\noindent \textbf{Prompt Variations:} 
In our evaluations, we aimed to control the impact of prompt design by minimizing unnecessary differences across different evaluation protocols. 
However, we acknowledge that a more thorough evaluation involving multiple prompt variants for each protocol would likely produce more stable results.

\noindent \textbf{Qualitative Human Evaluation:} We primarily used high-quality human annotation datasets for our quantitative meta-evaluation. 
Nevertheless, we recognize the lack of qualitative human evaluation, especially concerning the rationales generated by different LLM-evaluators, which could provide further insights into their limitations.
We provide a preliminary case study in Appendix~\ref{appx_case_study} showcasing the error patterns of the base LLMs, and another in Appendix~\ref{appx_case_study_llama3_70b} demonstrating the effect of different evaluation protocols on the same base LLM.

\section*{Acknowledgements} 
We are grateful to OpenAI's Researcher Access Program and Together AI for provision of LLM API credits.

\bibliography{anthology,custom}


\appendix

\section{Dataset Examples}
\label{appx_data_examples}

We randomly select two data examples from each of the four datasets in our study (\S\ref{sec:data_models}), and present them in Table~\ref{tab:dataset_ex_1} and Table~\ref{tab:dataset_ex_2}.

\section{Details of Base LLMs}
\label{appx_model_details}

We provide brief descriptions for the 38 base LLMs adopted in this study in Table~\ref{tab:appx_model_registry}, discussed in \S\ref{sec:status-quo}.

\section{Details of Evaluation Protocols}
\label{appx_prompts}

We provide a list of the 15 evaluation protocols investigated in this study in Table~\ref{tab:appx_method_registry}, detailed in \S\ref{sec:all_protocols}.
Below, we provide the prompt templates used for all prompting-based evaluation protocols in \S\ref{sec:all_protocols}.

\subsection{Prompt for Base Evaluation Protocol}
\label{appx_base_prompts}
Figure~\ref{fig:prompt_base} shows the prompt for the \texttt{base} protocol.
It corresponds to the \textit{Vanilla+Rules} prompting strategy proposed in \citet{{zeng2024evaluating}}.

\subsection{Prompts for Benchmark Evaluation Protocols}
\label{appx_benchmark_prompts}

We provide the prompts adopted from AlpacaEval~\cite{alpaca_eval} (Figure ~\ref{fig:prompt_alpaca}), ArenaHard~\cite{li2024crowdsourced} (Figure~ \ref{fig:prompt_arena}), and WildBench~\cite{lin2024wildbench}  (Figure~ \ref{fig:prompt_wildbench}).
The original evaluation protocols of ArenaHard and WildBench perform five-scale pairwise comparisons between the output pairs.
To better suit our task format, we modify their evaluation task to a binary pairwise comparison.
The evaluation protocol of WildBench requires a task-specific checklist of output quality to aid the evaluation.
In WildBench, these checklists were created using GPT-4-Turbo and Claude-3-Opus and manually reviewed.
Following a similar approach, we use GPT-4o to generate these checklists for the 4 datasets used in this work.

\subsection{Prompts for Enhanced Evaluation Protocols}
\label{appx_enhanced_prompts}

We list the prompts for \texttt{cot} (Figure~\ref{fig:prompt_cot}), \texttt{metric} (Figure~\ref{fig:prompt_metric_gen} \& Figure~\ref{fig:prompt_metric}), \texttt{reference} (Figure~\ref{fig:prompt_reference}), \texttt{metric+reference} (Figure~\ref{fig:prompt_metric_reference}), and \texttt{swap\&synthesize} (Figure~\ref{fig:prompt_swap_synthesis}).
These prompt templates are proposed by \citet{{zeng2024evaluating}}.

\subsection{Prompts for Complex Evaluation Protocols}
\label{appx_complex_prompts}
We present our prompts for \texttt{fine-grained-diff} (Figure~\ref{fig:prompt_finegrained_differences}), \texttt{multi-role-round2} (Figure~\ref{fig:prompt_multi-role_debate}), \texttt{multi-aspect-single} (Figure~\ref{fig:prompt_multi_aspect_one}), \texttt{multi-aspect-two} (Figure~\ref{fig:prompt_multi_aspect_two_analysis} and Figure~\ref{fig:prompt_multi_aspect_two_final}), \texttt{gpt4-reference} (Figure~\ref{fig:prompt_gpt4_reference}), and \texttt{prepair} (Figure~\ref{fig:prompt_prepair_pointwise} and Figure~\ref{fig:prompt_prepair_pairwise}).

\section{Case Study}
\label{appx_case_study}
We perform a qualitative analysis of the evaluation performance of \llama-3-70B and identify three main error patterns that impact its performance in various instances. The following paragraphs outline these error patterns, and we present a specific case study in Table~\ref{tab:appx_error_case}.

\paragraph{Surface-level deception \textsc{(surface)}} The model tends to favor outputs that appear more positive or have more structured presentations like numbered lists or professional layout despite clear disadvantages in addressing the instruction task compared to less structured but more appropriate and accurate responses. This failure mode is a recognized pattern across LLM-evaluators~\citep{zheng2024judging,wang-etal-2024-large-language-models-fair}.

\paragraph{Overlooking crucial context \textsc{(overlook)}} The model sometimes fails to identify important context in the question when it is not explicitly emphasized. This can lead to generic responses that miss key elements specific to the task at hand.

\paragraph{Counting issues \textsc{(counting)}} This applies to both numerical counting (for example, ``generate an 8-letter password'') and text-based context-relevant counting. The model often fails to count letters correctly or prefers longer outputs that exceed the instructed requirements, contrary to human annotators' preference for accurate answers.

\section{Generation Samples}
\label{appx_case_study_llama3_70b}
\begin{figure*}[t!]
  \centering
  \includegraphics[width=0.9\textwidth]{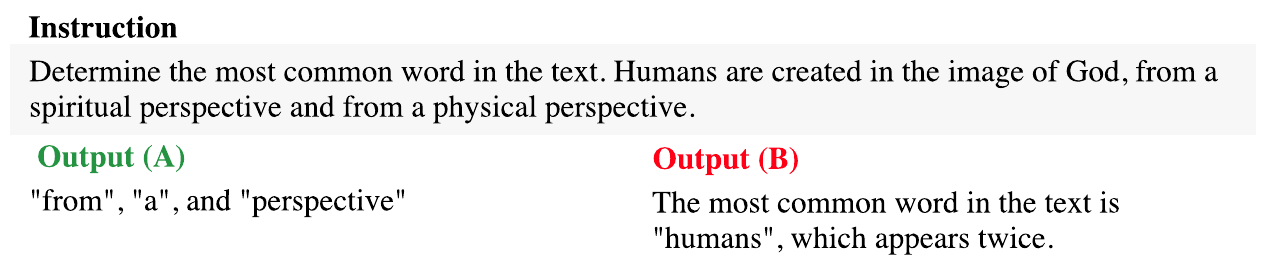}
  \caption{An instance from \llmbarnatural dataset. Output (A) is the gold preferred response.}
  \label{fig:appx_case_70b_data}
\end{figure*}

\begin{figure*}[t!]
  \centering
  \includegraphics[width=1\textwidth]{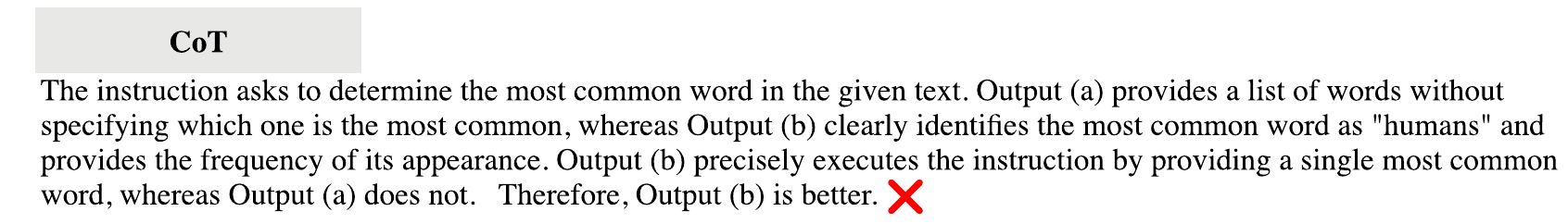}
  \caption{\llama-3-70b output under \texttt{cot} protocol.}
  \label{fig:appx_case_70b_cot}
\end{figure*}

\begin{figure*}[t!]
  \centering
  \includegraphics[width=1\textwidth]{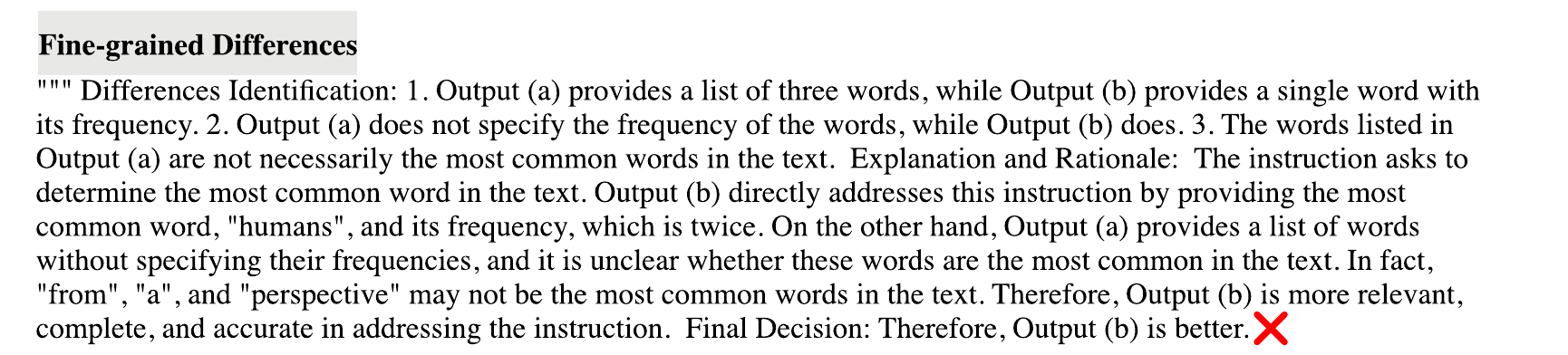}

  \caption{\llama-3-70b output under \texttt{fine-grained-diff} protocol.}
    \label{fig:appx_case_70b_fine}
\end{figure*}

\begin{figure*}[t!]
  \centering
  \includegraphics[width=1\textwidth]{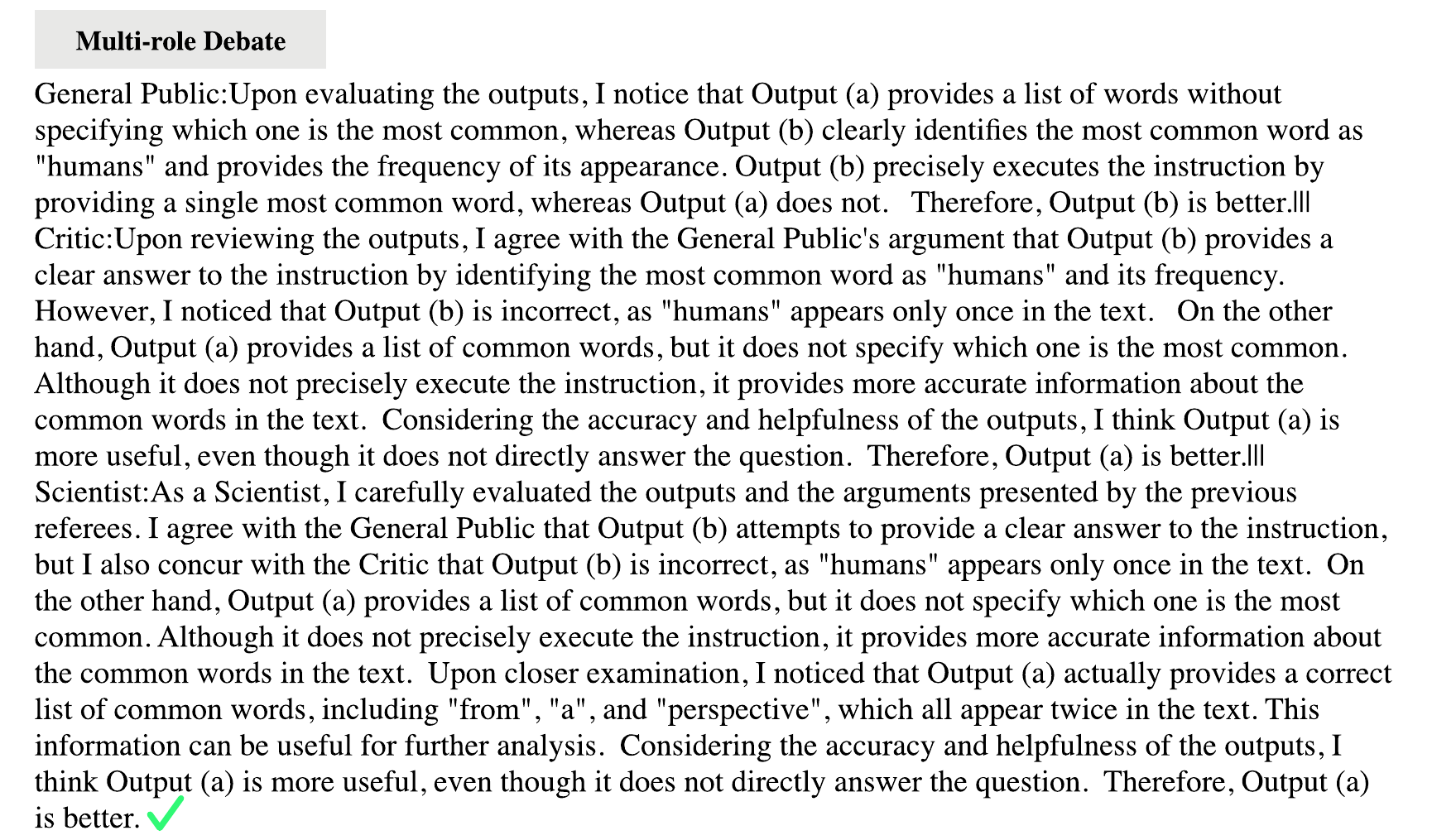}
  \caption{\llama-3-70b output under \texttt{multi-role-round2} protocol.}
  \label{fig:appx_case_70b_multi_role}
\end{figure*}

\begin{figure*}[t!]
  \centering
  \includegraphics[width=1\textwidth]{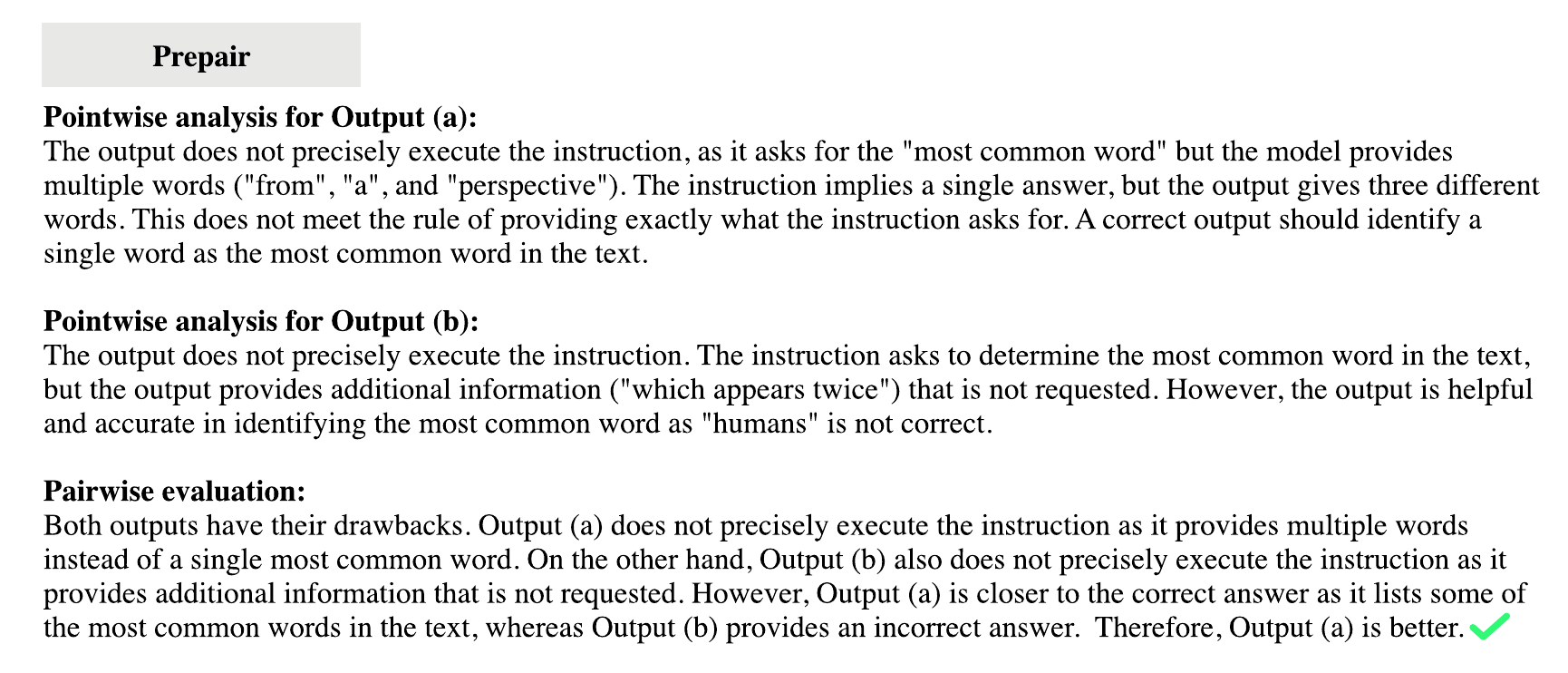}
  \caption{\llama-3-70b outputs under \texttt{prepair} protocol. Outputs from the pointwise analysis stage and the pairwise evaluation stage are presented.}
  \label{fig:appx_case_70b_prepair}

\end{figure*}

We present generation samples across different evaluation protocols of \llama-3-70B model for an example from \llmbarnatural, as shown in Figure~\ref{fig:appx_case_70b_data}. Model evaluations of \llama-3-70B under selected evaluation protocols are presented in Figure~\ref{fig:appx_case_70b_cot} (\texttt{cot}), Figure~\ref{fig:appx_case_70b_fine} (\texttt{fine-grained-diff}), Figure~\ref{fig:appx_case_70b_multi_role} (\texttt{multi-role-round2}),  and Figure~\ref{fig:appx_case_70b_prepair} (\texttt{prepair}).
Our analysis of the evaluation protocols reveals interesting disparities in their effectiveness. For example, the \texttt{cot} fails because it overlooks the factual error regarding the frequency of the word ``humans'' in one of the candidate outputs. In contrast, 
\texttt{multi-role-round2}'s success can be attributed to its multi-perspective debate approach, which allows for a more thorough examination and fact-checking of the claims made in each output.

%

%

\begin{table*}[ht]
  \centering
  \footnotesize
    \begin{tabular}{p{24.5em}c}
    \toprule
    \multicolumn{2}{l}{\textbf{\llmbarnatural}} \\
    \midrule
    \multicolumn{2}{l}{ Explain: Pathos, Ethos, Logos. } \\
    \midrule
    \multicolumn{1}{l}{\textbf{\lime{Output 1} }} & \multicolumn{1}{l}{\textbf{\red{Output 2} }} \\
   
Ethos is the ethical appeal, which is used to convince an audience of the author's credibility or character. 

Pathos is the emotional appeal, which is used to persuade an audience by appealing to their emotions. 

Logos is the appeal to logic and is used to convince an audience by the use of logic and reason.

& \multicolumn{1}{p{24.5em}}{Pathos is a tool of persuasion that is used to appeal to readers' emotions by arousing positive or negative feelings. 

It can be used in rhetoric, literature, film, and other forms of expression.

While pathos is used to draw an emotional response, the other rhetorical appeals appeal to credibility and logic, respectively.

Ethos is the distinguishing character, sentiment, moral nature, or guiding beliefs of a person, group, or institution. 

It is often evident as facts, numbers, or statistics and is used to convince an audience of an argument. } \\
    \midrule
    \multicolumn{2}{l}{\textbf{\adversarial}} \\
    \midrule
    \multicolumn{2}{l}{ Name an animal that is commonly kept as a pet.} \\
    \midrule
    \multicolumn{1}{l}{\textbf{\lime{Output 1}}} & \multicolumn{1}{l}{\textbf{\red{Output 2} }} \\
   
        Dogs are commonly kept as pets.
    & \multicolumn{1}{p{24.5em}}{Dogs, cats, and fish.} \\
    \midrule
    
    \multicolumn{2}{l}{\textbf{\mtbench}} \\
    \midrule
    \multicolumn{2}{l}{ A is the father of B. B is the father of C. What is the relationship between A and C?} \\
    \midrule
    \multicolumn{1}{l}{\textbf{\red{Output 1} }} & \multicolumn{1}{l}{\textbf{\lime{Output 2} }} \\
        A is the grandfather of C.
    & \multicolumn{1}{p{24.5em}}{

        * A is the father of B
        
        * B is the father of C
        
        * Therefore, A is the grandfather of C
        
        So the relationship between A and C is:
        
        Grandfather - Grandson

} \\

    \midrule
    \multicolumn{2}{l}{\textbf{\instrusum}} \\
    \midrule
\multicolumn{2}{l}{ 
    \parbox{13cm}{ 
         Here is an article:\\
        It is a book with 277 authors but with one shared experience - becoming a new mother in 2020.
        
        The mothers speak about their isolation, ...
        
        Please summarize the above article based on the given requirement:
    }
} \\
    \midrule
    \multicolumn{1}{l}{\textbf{\red{Output 1} }} & \multicolumn{1}{l}{\textbf{\lime{Output 2} }} \\

Born in Lockdown is a book written by 277 new mothers in the UK about their experiences of becoming a mother during the pandemic. The mothers share their feelings of isolation, uncertainty, and pressure on their mental health, but also highlight the silver linings and positives that have come out of this joint experience. The project was set up by Bristol novelist Emylia Hall to encourage writing for pleasure and purpose. The book is an unflinchingly honest and moving account of new motherhood during the pandemic, where resilience, love, and hope shine through. The book is available for free download from the Mothership Writers website, with voluntary donations going to the stillbirth and neonatal death charity Sands.
& \multicolumn{1}{p{24.5em}}{Mothers in 2020 have experienced a unique set of challenges due to the pandemic, including isolation, uncertainty, and pressure on their mental health. Despite the hardships, many have found silver linings and positives that have come out of the shared experience. Through the writing project Born in Lockdown, 277 mothers from across the UK have shared their stories, expressing a range of emotions including resilience, love, and hope. } \\
    \bottomrule
    \end{tabular}%
  \caption{Selected examples from our datasets (\S\ref{sec:data_models}). The green label indicates the output is the preferred output.}
  \label{tab:dataset_ex_1}
\end{table*}%

\begin{table*}[t!]
  \centering
  \footnotesize
    \begin{tabular}{p{24.5em}c}
    \toprule
    \multicolumn{2}{l}{\textbf{\llmbarnatural}} \\
    \midrule
    \multicolumn{2}{l}{ 
    \parbox{13cm}{Given the following input statement, provide the type of sentiment in the statement and the extent of sentiment in terms of sentiment scores. This restaurant has the best service I've ever seen. You can design the score scale by yourself.}
    
    } \\
    \midrule
    \multicolumn{1}{l}{\textbf{\red{Output 1} }} & \multicolumn{1}{l}{\textbf{\lime{Output 2} }} \\
   
    This statement has a sentiment score of 9 / 10

& \multicolumn{1}{p{24.5em}}{Positive sentiment, sentiment score: 9.5 out of 10 } \\
    \midrule
    \multicolumn{2}{l}{\textbf{\adversarial}} \\
    \midrule
    \multicolumn{2}{l}{ Analyze the results of the following test scores. 90, 86, 75, 63} \\
    \midrule
    \multicolumn{1}{l}{\textbf{\red{Output 1}}} & \multicolumn{1}{l}{\textbf{\lime{Output 2} }} \\
   
        The average grade is 78.75.
    & \multicolumn{1}{p{24.5em}}{The test scores show a decreasing trend, with the first score being the highest and the last score being the lowest. This could indicate that the test taker's knowledge and skills were increasingly weak as the test progressed.} \\
    \midrule
    
    \multicolumn{2}{l}{\textbf{\mtbench}} \\
    \midrule
    \multicolumn{2}{l}{ 
    \parbox{15cm}{Identify the countries, their capitals, and the languages spoken in the following sentences. Output in JSON format.
    
    a) Amidst the idyllic vistas, Copenhagen, Denmark's capital, captivates visitors with its thriving art scene and the enchanting Danish language spoken by its inhabitants.
    
    b) Within the enchanting realm of Eldoria, one discovers Avalore, a grandiose city that emanates an ethereal aura. Lumina, a melodious language, serves as the principal mode of communication within this mystical abode.
    
    c) Nestled amidst a harmonious blend of age-old customs and contemporary wonders, Buenos Aires, the capital of Argentina, stands as a bustling metropolis. It is a vibrant hub where the expressive Spanish language holds sway over the city's inhabitants.}

    } \\
    \midrule
    \multicolumn{1}{l}{\textbf{\lime{Output 1} }} & \multicolumn{1}{l}{\textbf{\red{Output 2} }} \\
\{``sentences'': [

\{
``text'': ``Amidst the idyllic vistas, Copenhagen, Denmark's capital, captivates visitors with its thriving art scene and the enchanting Danish language spoken by its inhabitants.'',

``country'': ``Denmark'',

``capital'': ``Copenhagen'',

``language'': ``Danish''
\},

\{
``text'': ''Within the enchanting realm of Eldoria, one discovers Avalore, a grandiose city that emanates an ethereal aura. Lumina, a melodious language, serves as the principal mode of communication within this mystical abode.'',

``country'': ``Eldoria (fictional)'',

``capital'': ``Avalore (fictional)'',

``language'': ``Lumina (fictional)''
\},

\{
``text'': ``Nestled amidst a harmonious blend of age-old customs and contemporary wonders, Buenos Aires, the capital of Argentina, stands as a bustling metropolis. It is a vibrant hub where the expressive Spanish language holds sway over the city's inhabitants.'',

``country'': ``Argentina'',

``capital'': ``Buenos Aires'',

``language'': ``Spanish''
\}

]\}
    & \multicolumn{1}{p{24.5em}}{

    a) The capital of Denmark is Copenhagen. The language spoken in Denmark is Danish.
    
    b) The capital of Eldoria is Avalore. The language spoken in Eldoria is Lumina.
    
    c) The capital of Argentina is Buenos Aires. The language spoken in Argentina is Spanish.
 
} \\
\midrule
    \multicolumn{2}{l}{\textbf{\instrusum}} \\
    \midrule
    \multicolumn{2}{l}{ 
    \parbox{15cm}{ 
    Here is an article:
    
    As pressure to tackle London's housing crisis grows, a number of controversial views are being offered on how the problem...
    
    Please summarize the above article based on the given requirement:
    }
    } \\
    \midrule
    \multicolumn{1}{l}{\textbf{\lime{Output 1} }} & \multicolumn{1}{l}{\textbf{\red{Output 2} }} \\
   
Research director at The Adam Smith Institute, Sam Bowman, suggests that building on London's green belt lands could provide bigger and cheaper homes for everyone. However, Ann Goddard of Elstree \& Borehamwood Greenbelt Society argues that green belt lands are important for recreation purposes and to maintain a separation between settlements. Bowman proposes a compromise by differentiating between areas of beauty and intensive farmland within the green belt.

& \multicolumn{1}{p{24.5em}}{One solution to London's housing crisis is to build more properties on green belt lands. This would mean bigger and cheaper homes for everyone, but there is opposition to this idea as green belt land is seen as important for recreation and to make a demarcation between settlements. A compromise could be reached by differentiating between areas of beauty and intensive farmland.} \\
    \bottomrule
    \end{tabular}%
  \caption{Selected examples from our datasets (\S\ref{sec:data_models}). The green label indicates the output is the preferred output.}
  \label{tab:dataset_ex_2}
  
\end{table*}%

\begin{table*}[ht!]
    \centering
    \footnotesize
    \begin{tabular}{@{}lclc@{}}
    \toprule
    \multicolumn{1}{l}{\textbf{Name}} & \textbf{Size} & \textbf{License} & \multicolumn{1}{c}{\textbf{Description}} \\ \midrule
    \href{https://huggingface.co/google/gemma-2b-it}{gemma-2b}       & 2b  & Gemma & \multirow{2}{7cm}{Gemma is a family of open models from Google \citep{team2024gemma}}                       \\
    \href{https://huggingface.co/google/gemma-7b-it}{gemma-7b}       & 7b  & Gemma &                                          \\ \midrule
    \href{https://huggingface.co/THUDM/glm-4-9b-chat}{glm-4-9b}       & 9b  & GLM-4 & \parbox{7cm}{GLM-4-9B is an open-source version of the latest generation of pre-trained models launched by Zhipu AI~\citep{du2022glm}.}                                        \\ \midrule
    \href{https://huggingface.co/01-ai/Yi-1.5-9B-Chat}{yi-1.5-9b}       & 9b  & Yi & \multirow{2}{7cm}{Yi series are bilingual language models trained on a 3T multilingual corpus by 01.AI \citep{ai2024yi}}                       \\
    \href{https://huggingface.co/01-ai/Yi-1.5-34B-Chat}{yi-1.5-34b}       & 34b & Yi &                                          \\ \midrule
    \href{https://huggingface.co/meta-llama/Llama-2-7b-chat-hf}{llama-2-7b}       & 7b  & \llama 2 Community & \multirow{3}{7cm}{\llama 2 models are the latest generation developed by Meta AI \citep{touvron2023llama}, pretrained on 2.2T tokens.}                       \\
    \href{https://huggingface.co/meta-llama/Llama-2-13b-chat-hf}{llama-2-13b}       & 13b & \llama 2 Community &                                          \\
    \href{https://huggingface.co/meta-llama/Llama-2-70b-chat-hf}{llama-2-70b}       & 70b & \llama 2 Community &                                          \\
    \href{https://huggingface.co/meta-llama/Meta-Llama-3-8B-Instruct}{llama-3-8b}       & 8b  & \llama 3 Community & \multirow{2}{7cm}{\llama 3 are the latest open models from Meta AI \citep{meta_llama_3}, pretrained on 15T tokens.}                       \\
    \href{https://huggingface.co/meta-llama/Meta-Llama-3-70B-Instruct}{llama-3-70b}       & 70b & \llama 3 Community &                                          \\ 
    \href{https://huggingface.co/meta-llama/Meta-Llama-3.1-8B-Instruct}{llama-3.1-8b}       & 8b  & \llama 3.1 Community & \multirow{3}{7cm}{\llama 3.1 collection offers a series of multilingual models that outperform many open and closed chat models on industry benchmarks \citep{dubey2024llama}.}                       \\
    \href{https://huggingface.co/meta-llama/Meta-Llama-3.1-70B-Instruct}{llama-3.1-70b}       & 70b & \llama 3.1 Community &                                          \\
    \href{https://huggingface.co/meta-llama/Meta-Llama-3.1-405B-Instruct}{llama-3.1-405b}       & 405b & \llama 3.1 Community &                                          \\ \midrule 
    \href{https://huggingface.co/allenai/tulu-2-7b}{\tulu-2-7b}       & 7b  & AI2 ImpACT Low-risk & \multirow{6}{7cm}{\tulu V2 \citep{ivison2023camels} is a series of \llama 2 based models that are instruction-tuned on \tulumix.}                       \\
    \href{https://huggingface.co/allenai/tulu-2-dpo-7b}{\tulu-2-dpo-7b}       & 7b  & AI2 ImpACT Low-risk &                                          \\
    \href{https://huggingface.co/allenai/tulu-2-13b}{\tulu-2-13b}       & 13b & AI2 ImpACT Low-risk &                                          \\
    \href{https://huggingface.co/allenai/tulu-2-dpo-13b}{\tulu-2-dpo-13b}       & 13b & AI2 ImpACT Low-risk &                                          \\
    \href{https://huggingface.co/allenai/tulu-2-70b}{\tulu-2-70b}       & 70b & AI2 ImpACT Low-risk &                                          \\
    \href{https://huggingface.co/allenai/tulu-2-dpo-70b}{\tulu-2-dpo-70b}       & 70b & AI2 ImpACT Low-risk &                                          \\ \midrule
    \href{https://deepmind.google/technologies/gemini/}{gemini-1.0-pro}       & -             & Proprietary & \multirow{3}{7cm}{Gemini models are the most capable multimodal models from Google featuring long context lengths \citep{team2023gemini}.}                       \\
    \href{https://deepmind.google/technologies/gemini/}{gemini-1.5-flash}       & -             & Proprietary &                                          \\
    \href{https://deepmind.google/technologies/gemini/}{gemini-1.5-pro}       & -             & Proprietary &                                          \\ \midrule
    \href{https://huggingface.co/Qwen/Qwen1.5-32B-Chat}{qwen-1.5-32b}       & 32b  & Qianwen & \multirow{4}{7cm}{Qwen is a family of models built by Alibaba Cloud~\citep{qwen}. Qwen1.5 and Qwen2 have recently surpassed most open models on common benchmarks.}                       \\
    \href{https://huggingface.co/Qwen/Qwen1.5-72B-Chat}{qwen-1.5-72b}       & 72b  & Qianwen &                                          \\
    \href{https://huggingface.co/Qwen/Qwen2-72B-Instruct}{qwen-2-72b}       & 72b  & Qianwen &                                          \\
    \href{https://huggingface.co/Qwen/Qwen2.5-72B-Instruct}{qwen-2.5-72b}       & 72b  & Qianwen &                                          \\
        \midrule
    \href{https://huggingface.co/mistralai/Mistral-7B-Instruct-v0.3}{mistral-7b-v0.3}          & 7b  & Apache 2.0 & \multirow{2}{7cm}{Instruction-tuned versions of Mistral models~\citep{jiang2023mistral} from Mistral AI.}                       \\
    \href{https://mistral.ai/news/mistral-large/}{mistral-large}       & -             & Proprietary &                                          \\ \midrule
    \href{https://huggingface.co/mistralai/Mixtral-8x7B-Instruct-v0.1}{mixtral-8x7b}       & 8x7b & Apache 2.0 & \parbox{7cm}{Mixtral-8x22B is a pretrained generative Sparse Mixture of Experts (MoE) from Mistral AI~\citep{jiang2024mixtral}}                                        \\ \midrule
    \href{https://www.anthropic.com/api}{claude-3-haiku}       & -             & Proprietary & \multirow{3}{7cm}{Claude-3-Haiku and Claude 3 Opus, and Claude 3.5 Sonnet are top proprietary models trained by Anthropic PBC \citep{claude}.}                       \\
    \href{https://www.anthropic.com/api}{claude-3-opus}       & -             & Proprietary &                                          \\
     \href{https://www.anthropic.com/api}{claude-3.5-sonnet}       & -             & Proprietary &                                          \\
        \midrule
    \href{https://platform.openai.com/docs/models}{gpt-3.5-turbo-0125}       & -             & Proprietary & \multirow{3}{7cm}{GPT models are strong proprietary models~\citep{achiam2023gpt} from OpenAI. ``o1'' model was released in September 2024 with strong reasoning capability. }                       \\
             \href{https://platform.openai.com/docs/models}{gpt-4-0613}                   & -             & Proprietary & \\

    \href{https://platform.openai.com/docs/models}{gpt-4o-2024-05-13}                   & -             & Proprietary & \\
     \href{https://platform.openai.com/docs/models}{gpt-4o-2024-08-06}                   & -             & Proprietary & \\
  \href{https://platform.openai.com/docs/models}{o1-mini-2024-0912}                   & -             & Proprietary & \\
 \midrule
    \href{https://huggingface.co/prometheus-eval/prometheus-8x7b-v2.0}{prometheus-2-8x7b}       & 8x7b  & Apache 2.0 & \parbox{7cm}{Prometheus 2 is an alternative to GPT-4 for fine-grained evaluation of LLMs and reward models for RLHF, based on Mistral-Instruct~\citep{kim2024prometheus2}}                                        \\ 
    \href{https://huggingface.co/NCSOFT/Llama-3-OffsetBias-8B}{offsetbias-lm}       & 8b  & \llama 3 Community & \parbox{7cm}{OffsetBias is a generative judge model for pairwise preference evaluation, designed to be robust against various evaluation biases~\citep{park2024offsetbias}}                                        \\ \midrule
    
  \href{https://huggingface.co/nvidia/Nemotron-4-340B-Reward}{nemotron-4-340b-reward}       & 340b  & NVIDIA Open Model & \parbox{7cm}{A multi-aspect reward model for synthetic data generation and RLAIF, based on Nemotron-4-340B-Base~\citep{wang2024helpsteer2}}   \\                                
    \href{https://huggingface.co/NCSOFT/Llama-3-OffsetBias-RM-8B}{offsetbias-rm}       & 8b  & \llama 3 Community & \parbox{7cm}{Reward model trained on OffsetBias dataset, designed to be robust against various evaluation biases~\citep{park2024offsetbias}}                                        \\ 

     \bottomrule
    \end{tabular}
    \caption{Model registry and metadata in our study used in \S\ref{sec:status-quo}.}
    \label{tab:appx_model_registry}
    \end{table*}
\begin{table*}[ht!]
    \centering
    \footnotesize
    \begin{tabular}{@{}lll@{}}
    \toprule
    \textbf{Full Name} & \textbf{Code Name} & \textbf{Related Work} \\
    \midrule
    Base Pairwise Evaluation & base & \citet{zeng2024evaluating} \\
    Chain-of-Thought & cot & \citet{zeng2024evaluating,wei2022chain} \\
    Metric Generation & metric & \citet{zeng2024evaluating} \\
    Reference Generation & reference & \citet{zeng2024evaluating}  \\
    Metric and Reference & metric+reference & \citet{zeng2024evaluating} \\
    Swap and Synthesize & swap\&synthesize  & \citet{zeng2024evaluating} \\
    Fine-grained Differences & fine-grained-diff & Inspired by \citet{liu-etal-2023-revisiting, liu-etal-2023-towards-interpretable, min-etal-2023-factscore} \\
    Multi-Role Debate (Single Round) & multi-role-round1  & \citet{chan2024chateval} \\
    Multi-Role Debate (Two Rounds) & multi-role-round2  & \citet{chan2024chateval} \\
    Multi-Aspect Comparison (Two-Stage) & multi-aspect-two  & \citet{li2024decompose} \\
    Multi-Aspect Comparison (Single-Stage) & multi-aspect-single  & \citet{li2024decompose} \\
    GPT-4 Reference & gpt4-reference & Modified from \citet{zeng2024evaluating} \\
    Pointwise Reasoning Enhanced Pairwise & prepair  & \citet{jeong2024prepair} \\
    Self-Consistency over CoTs & cot\&consistency & \citet{wang2023selfconsistency} \\
    Self-Consistency over Different Protocols & protocol-consistency & \citet{wang2023selfconsistency} \\
    \bottomrule
    \end{tabular}
    \caption{Method registry for evaluation protocols in \S\ref{sec:all_protocols}}
    \label{tab:appx_method_registry}
\end{table*}
\clearpage
\begin{figure*}[t!]
\begin{tcolorbox}[colback=black!3!white, colframe=black!70!white, title=Base, fontupper=\footnotesize, fonttitle=\footnotesize]
\textbf{[System Message]} \\
You are a helpful assistant in evaluating the quality of the outputs for a given instruction. Your goal is to select the best output for the given instruction.
\newline
\newline
\textbf{[User Message]}\\
Select the Output (a) or Output (b) that is better for the given instruction. The two outputs are generated by two different AI chatbots respectively. \\
\newline
Here are some rules of the evaluation: \\
(1) You should prioritize evaluating whether the output honestly/precisely/closely executes the instruction, then consider its helpfulness, accuracy, level of detail, harmlessness, etc. \\
(2) Outputs should NOT contain more/less than what the instruction asks for, as such outputs do NOT precisely execute the instruction.\\
(3) You should avoid any potential bias and your judgment should be as objective as possible. For example, the order in which the outputs were presented should NOT affect your judgment, as Output (a) and Output (b) are equally likely to be the better.
\newline
\newline
Do NOT provide any explanation for your choice.\\
Do NOT say both / neither are good.\\
You should answer using ONLY "Output (a)" or "Output (b)". Do NOT output any other words.\\
\newline

\# Instruction: \\
\{INSTRUCTION\}
\newline
\newline
\# Output (a): \\
\{OUTPUT\_1\}
\newline
\newline
\# Output (b): \\
\{OUTPUT\_2\}
\newline
\newline
\# Which is better, Output (a) or Output (b)? Your response should be either "Output (a)" or "Output (b)":
\end{tcolorbox}
\caption{Prompt for \texttt{base} protocol described in \S\ref{sec:all_protocols}.}
\label{fig:prompt_base}
\end{figure*}

\begin{figure*}[t!]
\begin{tcolorbox}[colback=black!3!white, colframe=black!70!white, title=AlpacaEval, fontupper=\footnotesize, fonttitle=\footnotesize]
\textbf{[System Message]} \\
You are a highly efficient assistant, who evaluates and selects the best large language model (LLMs) based on the quality of their responses to a given instruction. This process will be used to create a leaderboard reflecting the most accurate and human-preferred answers.
\newline
\newline
\textbf{[User Message]}\\
I require a leaderboard for various large language models. I'll provide you with prompts given to these models and their corresponding outputs. Your task is to assess these responses, and select the model that produces the best output from a human perspective. \\
\newline

\#\# Instruction

\{ \\
    "instruction": """\{INSTRUCTION\}""",\\
\}\\
\newline
\newline
\#\# Model Outputs
\newline

Here are the unordered outputs from the models. Each output is associated with a specific model, identified by a unique model identifier.
\\
\{\\
\hspace*{1cm}  \{
        "model\_identifier":  "m",\\
         \hspace*{1cm} "output": """\{OUTPUT\_1\}"""\\
   \hspace*{1cm}  \},\\
    \hspace*{1cm}\{\\
       \hspace*{1cm} "model\_identifier": "M",\\
        \hspace*{1cm}"output": """\{OUTPUT\_2\}"""\\
   \hspace*{1cm}       \}\\
\}\\
\newline
\#\# Task
\newline
Evaluate the models based on the quality and relevance of their outputs, and select the model that generated the best output. Answer by providing the model identifier of the best model. We will use your output as the name of the best model, so make sure your output only contains one of the following model identifiers and nothing else (no quotes, no spaces, no new lines, ...): Model (m) or Model (M).
\\ 

\#\# Which is better, Model (m) or Model (M)? Your response should be either "Model (m)" or "Model (M)":
\end{tcolorbox}
\caption{Prompt for AlpacaEval baseline described in \S\ref{baselines_protocols}}
\label{fig:prompt_alpaca}
\end{figure*}

\begin{figure*}[t!]
\begin{tcolorbox}[colback=black!3!white, colframe=black!70!white, title=ArenaHard, fontupper=\footnotesize, fonttitle=\footnotesize]
\textbf{[System Message]} \\
Please act as an impartial judge and evaluate the quality of the responses provided by two AI assistants to the user prompt displayed below. You will be given assistant A's answer and assistant B's answer. Your job is to evaluate which assistant's answer is better.
\newline
\textbf{[User Message]}\\
Begin your evaluation by generating your own answer to the prompt. You must provide your answers before judging any answers.
\newline

When evaluating the assistants' answers, compare both assistants' answers with your answer. You must identify and correct any mistakes or inaccurate information.
\newline

Then consider if the assistant's answers are helpful, relevant, and concise. Helpful means the answer correctly responds to the prompt or follows the instructions. Note when user prompt has any ambiguity or more than one interpretation, it is more helpful and appropriate to ask for clarifications or more information from the user than providing an answer based on assumptions. Relevant means all parts of the response closely connect or are appropriate to what is being asked. Concise means the response is clear and not verbose or excessive.
\newline

Then consider the creativity and novelty of the assistant's answers when needed. Finally, identify any missing important information in the assistants' answers that would be beneficial to include when responding to the user prompt.
\newline

After providing your explanation, you must always end your response with either "Therefore, Answer (a) is better." or "Therefore, Answer (b) is better." verbatim.
\newline
\newline

<|User Prompt|>\\
\{INSTRUCTION\}
\\ 
\\
<|The Start of Answer (a)|> \\
\{OUTPUT\_1\} \\
<|The End of Answer (a)|>
\\
\\
<|The Start of Answer (b)|> \\
\{OUTPUT\_2\}\\
<|The End of Answer(b)|>\\
\\

\# Decision (Give an explanation of your evaluation followed by either "Therefore, Answer (a) is better." or "Therefore, Answer (b) is better." verbatim. Always claim which is better at the end.):

\end{tcolorbox}
\caption{Prompt for ArenaHard described in \S\ref{baselines_protocols}}
\label{fig:prompt_arena}
\end{figure*}

\begin{figure*}[t!]
\begin{tcolorbox}[colback=black!3!white, colframe=black!70!white, title=WildBench, fontupper=\footnotesize, fonttitle=\footnotesize]
\textbf{[System Message]} \\
You are an expert evaluator. Your task is to evaluate the quality of the responses generated by two AI models. 
\newline
\textbf{[User Message]}\\
\# Instruction 
\\ \\ 
We will provide you with the user query and a pair of AI-generated responses (Response A and Response B). \\
You should first read the user query and the conversation history carefully for analyzing the task, and then evaluate the quality of the responses based on and rules provided below. \\
\newline

\# Conversation between User and AI
\\ 
\#\# User Query \\
<|begin\_of\_query|> \\
\{INSTRUCTION\} \\
<|end\_of\_query|>
\\ \\
\#\# Response A \\
<|begin\_of\_response\_A|> \\
\{OUTPUT\_1\}
\newline
<|end\_of\_response\_A|> \\

\#\# Response B \\
<|begin\_of\_response\_B|> \\
\{OUTPUT\_2\}
\\
<|end\_of\_response\_B|>
\\
\\
\# Evaluation   
\\ 
\#\# Checklist 
\newline
<|begin\_of\_checklist|> \\
\{CHECKLIST\} \\
<|end\_of\_checklist|> \\
\newline

Please use this checklist to guide your evaluation, but do not limit your assessment to the checklist.
\\
\\ 

\#\# Rules 
\newline
\newline
You should compare the above two responses based on your analysis of the user query. \\
You should first write down your analysis and the checklist that you used for the evaluation, and then provide your assessment according to the checklist. \\
You should always end your response by selecting the better response. \\
\newline

\#\# Output Format \\

First, please output your analysis for each model response, and then summarize your assessment to three aspects: "reason A=B", "reason A>B", and "reason B>A", and finally make your choice for the final assessment by selecting the better response (ties are NOT allowed). \\
\newline
Please provide your evaluation results in the following json format by filling in the placeholders in []: \\

``` \\
\{ \\
\hspace*{1cm} "analysis of A": "[analysis of Response A]", \\
  \hspace*{1cm}  "analysis of B": "[analysis of Response B]",\\
  \hspace*{1cm}  "reason of A=B": "[where Response A and B perform equally well]", \\
  \hspace*{1cm}  "reason of A>B": "[where Response A is better than Response B]", \\
  \hspace*{1cm}  "reason of B>A": "[where Response B is better than Response A]", \\
  \hspace*{1cm}  "choice": "[Response (A) or Response (B)]", \\
\} \\
```

\end{tcolorbox}
\caption{Prompt for WildBench baseline described in \S\ref{baselines_protocols}}
\label{fig:prompt_wildbench}
\end{figure*}

\begin{figure*}[t!]
\begin{tcolorbox}[colback=black!3!white, colframe=black!70!white, title=CoT, fontupper=\footnotesize, fonttitle=\footnotesize]
\textbf{[System Message]} \\
You are a helpful assistant in evaluating the quality of the outputs for a given instruction. Your goal is to select the best output for the given instruction.
\newline
\newline
\textbf{[User Message]}\\
After giving a brief explanation, select the Output (a) or Output (b) that is better for the given instruction. The two outputs are generated by two different AI chatbots respectively.
\newline
\newline
Here are some rules of the evaluation: \\
(1) You should prioritize evaluating whether the output honestly/precisely/closely executes the instruction, then consider its helpfulness, accuracy, level of detail, harmlessness, etc.\\
(2) Outputs should NOT contain more/less than what the instruction asks for, as such outputs do NOT precisely execute the instruction. \\
(3) You should avoid any potential bias and your judgment should be as objective as possible. For example, the order in which the outputs were presented should NOT affect your judgment, as Output (a) and Output (b) are equally likely to be the better.\\
\newline
\newline
You should first provide a brief explanation of your evaluation, and then always end your response with either "Therefore, Output (a) is better." or "Therefore, Output (b) is better." verbatim.\\
Do NOT say both / neither are good.\\
Do NOT output any other words.\\
Do NOT say "Output (a) is better" or "Output (b) is better" at the beginning. You should do reasoning and thinking before claiming which is better.\\
\newline
\# Instruction: \\
\{INSTRUCTION\}
\newline
\newline
\# Output (a): \\
\{OUTPUT\_1\}
\newline
\newline
\# Output (b): \\
\{OUTPUT\_2\}
\newline
\newline
\# Decision (Give a brief explanation of your evaluation followed by either "Therefore, Output (a) is better." or "Therefore, Output (b) is better." verbatim. Always claim which is better at the end. In your explanation, you should always use "Output (a)" or "Output (b)" to refer to the two outputs respectively.):
\end{tcolorbox}
\caption{Prompt for \texttt{cot} protocol described in \S\ref{sec:all_protocols}.}
\label{fig:prompt_cot}
\end{figure*}

\begin{figure*}[t!]
\begin{tcolorbox}[colback=black!3!white, colframe=black!70!white, title=Metric (metric generation prompt), fontupper=\footnotesize, fonttitle=\footnotesize]
\textbf{[System Message]} \\
You are a helpful assistant.
\newline
\newline
\textbf{[User Message]}\\
Please propose at most three concise questions about whether a potential output is a good output for a given instruction. Another assistant will evaluate different aspects of the output by answering all the questions.
\newline
\newline
Here are some rules of the evaluation: \\
(1) You should prioritize evaluating whether the output honestly/precisely/closely executes the instruction.
\\
(2) Outputs should NOT contain more/less than what the instruction asks for, as such outputs do NOT precisely execute the instruction.
\\
\newline
\# Instruction: \\
\{INSTRUCTION\}
\newline
\newline
\# Requirements for Your Output: \\
(1) The questions should **specifically** target the given instruction instead of some general standards, so the questions may revolve around key points of the instruction. \\
(2) You should directly give the questions without any other words. \\
(3) Questions are presented from most important to least important.\\
\newline
\# Please give your questions here:

\end{tcolorbox}
\caption{Prompt for metric generation stage of the \texttt{metric} protocol described in \S\ref{sec:all_protocols}.}
\label{fig:prompt_metric_gen}
\end{figure*}

\begin{figure*}[t!]
\begin{tcolorbox}[colback=black!3!white, colframe=black!70!white, title=Metric, fontupper=\footnotesize, fonttitle=\footnotesize]
\textbf{[System Message]} \\
You are a helpful assistant in evaluating the quality of the outputs for a given instruction. Your goal is to select the best output for the given instruction.
\newline
\newline
\textbf{[User Message]}\\
Select the Output (a) or Output (b) that is better for the given instruction. The two outputs are generated by two different AI chatbots respectively.
\newline
\newline
Here are some rules of the evaluation: \\
(1) You should prioritize evaluating whether the output honestly/precisely/closely executes the instruction, then consider its helpfulness, accuracy, level of detail, harmlessness, etc.\\
(2) Outputs should NOT contain more/less than what the instruction asks for, as such outputs do NOT precisely execute the instruction.\\
(3) You should avoid any potential bias and your judgment should be as objective as possible. For example, the order in which the outputs were presented should NOT affect your judgment, as Output (a) and Output (b) are equally likely to be the better.\\
\newline
\newline
Do NOT provide any explanation for your choice. \\
Do NOT say both / neither are good.\\
You should answer using ONLY "Output (a)" or "Output (b)". Do NOT output any other words.\\
\newline
\newline
\# Instruction: \\
\{INSTRUCTION\}
\newline
\newline
\# Output (a): \\
\{OUTPUT\_1\}
\newline
\newline
\# Output (b): \\
\{OUTPUT\_2\}
\newline
\newline
\# Questions about Outputs:\\
Here are at most three questions about the outputs, which are presented from most important to least important. You can do the evaluation based on thinking about all the questions. \\
\{QUESTIONS\}
\newline
\newline
\# Which is better, Output (a) or Output (b)? Your response should be either "Output (a)" or "Output (b)":
\end{tcolorbox}
\caption{Prompt for \texttt{metric} protocol described in \S\ref{sec:all_protocols}.}
\label{fig:prompt_metric}
\end{figure*}

\begin{figure*}[t!]
\begin{tcolorbox}[colback=black!3!white, colframe=black!70!white, title=Reference, fontupper=\footnotesize, fonttitle=\footnotesize]
\textbf{[System Message]} \\
You are a helpful assistant in evaluating the quality of the outputs for a given instruction. Your goal is to select the best output for the given instruction.
\newline
\newline
\textbf{[User Message]}\\
Select the Output (a) or Output (b) that is better for the given instruction. The two outputs are generated by two different AI chatbots respectively.
\newline\newline
Here are some rules of the evaluation:
\newline
(1) You should prioritize evaluating whether the output honestly/precisely/closely executes the instruction, then consider its helpfulness, accuracy, level of detail, harmlessness, etc.\\
(2) Outputs should NOT contain more/less than what the instruction asks for, as such outputs do NOT precisely execute the instruction.\\
(3) You should avoid any potential bias and your judgment should be as objective as possible. For example, the order in which the outputs were presented should NOT affect your judgment, as Output (a) and Output (b) are equally likely to be the better.\\
\newline
\newline
Do NOT provide any explanation for your choice. \\
Do NOT say both / neither are good.\\
You should answer using ONLY "Output (a)" or "Output (b)". Do NOT output any other words.\\
\newline
\newline
\# Instruction: \\
\{INSTRUCTION\}
\newline\newline
\# Output (a): \\
\{OUTPUT\_1\}
\newline\newline
\# Output (b):\\
\{OUTPUT\_2\}
\newline\newline
\# A reference output generated by a strong AI assistant: \\
\{REFERENCE\}
\newline
\newline
\# Which is better, Output (a) or Output (b)? Your response should be either "Output (a)" or "Output (b)":
\end{tcolorbox}
\caption{Prompt for \texttt{reference} protocol described in \S\ref{sec:all_protocols}.}
\label{fig:prompt_reference}
\end{figure*}

\begin{figure*}[t!]
\begin{tcolorbox}[colback=black!3!white, colframe=black!70!white, title=Metric + Reference, fontupper=\footnotesize, fonttitle=\footnotesize]
\textbf{[System Message]} \\
You are a helpful assistant in evaluating the quality of the outputs for a given instruction. Your goal is to select the best output for the given instruction.
\newline

\textbf{[User Message]}\\
Select the Output (a) or Output (b) that is better for the given instruction. The two outputs are generated by two different AI chatbots respectively.
\newline

Here are some rules of the evaluation:

(1) You should prioritize evaluating whether the output honestly/precisely/closely executes the instruction, then consider its helpfulness, accuracy, level of detail, harmlessness, etc.

(2) Outputs should NOT contain more/less than what the instruction asks for, as such outputs do NOT precisely execute the instruction.

(3) You should avoid any potential bias and your judgment should be as objective as possible. For example, the order in which the outputs were presented should NOT affect your judgment, as Output (a) and Output (b) are equally likely to be the better.
\newline

Do NOT provide any explanation for your choice. \\
Do NOT say both / neither are good. \\
You should answer using ONLY "Output (a)" or "Output (b)". Do NOT output any other words.\\

\# Instruction:

\{INSTRUCTION\}
\newline

\# Output (a):

\{OUTPUT\_1\}
\newline

\# Output (b):

\{OUTPUT\_2\}
\newline

\# Questions about Outputs:

Here are at most three questions about the outputs, which are presented from most important to least important. You can do the evaluation based on thinking about all the questions.

\{QUESTIONS\}
\newline

\# A reference output generated by a strong AI assistant:

\{REFERENCE\}
\newline

\# Which is better, Output (a) or Output (b)? Your response should be either "Output (a)" or "Output (b)":
\end{tcolorbox}
\caption{Prompt for \texttt{metric + reference} protocol described in \S\ref{sec:all_protocols}.}
\label{fig:prompt_metric_reference}
\end{figure*}

\begin{figure*}[t!]
\begin{tcolorbox}[colback=black!3!white, colframe=black!70!white, title=Swap\&Synthesize, fontupper=\footnotesize, fonttitle=\footnotesize]
\textbf{[System Message]} \\
You are a helpful assistant who reviews a debate between two other assistants in evaluating the quality of the outputs for a given instruction.
\newline

\textbf{[User Message]}\\
The two assistants, Assistant (a) and Assistant (b), are given an instruction, Output (a) and Output (b). They are asked to select the Output (a) or Output (b) that is better for the given instruction. Output (a) and Output (b) are generated by two different AI chatbots respectively.
\newline

Assistant (a) and Assistant (b) have conflicting evaluations. Your goal is to review their evaluations and give your final decision on which output is better.
\newline

Here are some rules of the evaluation:

(1) You should prioritize evaluating whether the output honestly/precisely/closely executes the instruction, then consider its helpfulness, accuracy, level of detail, harmlessness, etc.

(2) Outputs should NOT contain more/less than what the instruction asks for, as such outputs do NOT precisely execute the instruction.

(3) You should avoid any potential bias and your judgment should be as objective as possible. For example, the order in which the outputs were presented should NOT affect your judgment, as Output (a) and Output (b) are equally likely to be the better.
\newline

Now carefully review the instruction, Output (a), Output (b), and the debate between Assistant (a) and Assistant (b). Select the Output (a) or Output (b) that is better for the given instruction.

Do NOT provide any explanation for your choice.

Do NOT say both / neither are good.

You should answer using ONLY "Output (a)" or "Output (b)". Do NOT output any other words.
\newline

\# Instruction:

\{INSTRUCTION\}\newline

\# Output (b):

\{OUTPUT\_2\}\newline

\# Output (a):

\{OUTPUT\_1\}\newline

\# Debate between Assistant (a) and Assistant (b): 

\#\# Evaluation given by Assistant (a), who thinks Output (a) is better:

\{EXPLANATION\_1\}

\#\# Evaluation given by Assistant (b), who thinks Output (b) is better:

\{EXPLANATION\_2\}
\newline

\# Which is better, Output (a) or Output (b)? Your response should be either "Output (a)" or "Output (b)":
\end{tcolorbox}
\caption{Prompt for \texttt{swap\&synthesize} protocol described in \S\ref{sec:all_protocols}.}
\label{fig:prompt_swap_synthesis}
\end{figure*}

\begin{figure*}[t!]
\begin{tcolorbox}[colback=black!3!white, colframe=black!70!white, title=Fine-grained-diff, fontupper=\footnotesize, fonttitle=\footnotesize]
\textbf{[System Message]} \\
You are a helpful assistant in evaluating the quality of the outputs for a given instruction. Your goal is to select the best output for the given instruction.
\newline
\newline
\textbf{[User Message]}\\
After giving a detailed explanation, select either Output (a) or Output (b) as the better response for the given instruction. The two outputs are generated by two different AI chatbots respectively.
\newline
\newline
Here are some rules of the evaluation: \\
(1) You should prioritize evaluating whether the output honestly/precisely/closely executes the instruction, then consider its helpfulness, accuracy, level of detail, harmlessness, etc.\\
(2) Outputs should NOT contain more/less than what the instruction asks for, as such outputs do NOT precisely execute the instruction.\\
(3) You should avoid any potential bias and your judgment should be as objective as possible. For example, the order in which the outputs were presented should NOT affect your judgment, as Output (a) and Output (b) are equally likely to be the better.\\
\newline
You should first provide a detailed explanation of your evaluation, and then always end your response with either "Therefore, Output (a) is better." or "Therefore, Output (b) is better." verbatim.\\
Do NOT say both / neither are good.\\
Do NOT output any other words.\\
Do NOT say "Output (a) is better" or "Output (b) is better" at the beginning. You should do reasoning and thinking before claiming which is better.\\
\newline
Here is the evaluation plan:

1. Differences Identification: Enumerate the key fine-grained content differences observed between Output (a) and Output (b).\\
2. Explanation and Rationale: Provide explanations and rationale for which output better addresses the instruction by considering these differences, as well as other relevant factors such as relevance, completeness, coherence, and clarity.\\
3. Final Decision: End your response with either "Therefore, Output (a) is better." or "Therefore, Output (b) is better." verbatim.\\
\newline
Provide your response in the following format:\\
"""\\
Differences Identification: \\
\newline
1. [Difference 1] \\
2. [Difference 2] \\
... \\
N. [Difference N] \\
\newline
Explanation and Rationale: [Detailed explanation and rationale for your decision] \\

Final Decision: Therefore, Output (a)/Output (b) is better. \\
"""\\

\# Instruction: \\
\{INSTRUCTION\}
\newline
\newline
\# Output (a): \\
\{OUTPUT\_1\}
\newline
\newline
\# Output (b): \\
\{OUTPUT\_2\}
\newline

\# Your Response (Give a detailed explanation of your evaluation followed by either "Therefore, Output (a) is better." or "Therefore, Output (b) is better." verbatim. Always claim which is better at the end. In your explanation, you should always use "Output (a)" or "Output (b)" to refer to the two outputs respectively.):

\end{tcolorbox}
\caption{Prompt for \texttt{fine-graine-diff} protocol described in \S\ref{sec:all_protocols}.}
\label{fig:prompt_finegrained_differences}
\end{figure*}

\begin{figure*}[t!]
\begin{tcolorbox}[colback=black!3!white, colframe=black!70!white, title=Multi-role-round2, fontupper=\footnotesize, fonttitle=\footnotesize]
\textbf{[System Message]} \\
You are a helpful assistant who evaluates the quality of the outputs for a given instruction. Your goal is to select the best output for the given instruction.
\newline

\textbf{[User Message]}\\
Select the Output (a) or Output (b) that is better for the given instruction. The two outputs are generated by two different AI chatbots respectively.
\newline

Here are some rules of the evaluation:

(1) You should prioritize evaluating whether the output honestly/precisely/closely executes the instruction, then consider its helpfulness, accuracy, level of detail, harmlessness, etc.

(2) Outputs should NOT contain more/less than what the instruction asks for, as such outputs do NOT precisely execute the instruction.

(3) You should avoid any potential bias and your judgment should be as objective as possible. For example, the order in which the outputs were presented should NOT affect your judgment, as Output (a) and Output (b) are equally likely to be the better.
\newline

You should first provide a brief explanation of your evaluation, and then always end your response with either "Therefore, Output (a) is better." or "Therefore, Output (b) is better." verbatim.

Do NOT say both / neither are good.

Do NOT output any other words.

Do NOT say "Output (a) is better" or "Output (b) is better" at the beginning. You should do reasoning and thinking before claiming which is better.
\newline

There are a few other referees assigned the same task, it's your responsibility to discuss with them and think critically before you make your final judgement. You should avoid any potential bias and ensure that the order in which the responses were presented does not affect your judgment. Debate with others.

Always end your response with "Therefore, Output (a) is better." or "Therefore, Output (b) is better." verbatim. Make sure to make the claim to end your response.
\newline

\# Instruction: \\
\{INSTRUCTION\}
\newline

\# Output (a): \\
\{OUTPUT\_1\}
\newline

\# Output (b): \\
\{OUTPUT\_2\}
\newline

\# Previous referees' arguments: \\
\{CHAT\_HISTORY\}
\newline

\# Your role:

\{ROLE\_DESCRIPTION\}
\newline

\# Decision (Give a brief explanation of your evaluation followed by either "Therefore, Output (a) is better." or "Therefore, Output (b) is better." verbatim. Always claim which is better at the end. In your explanation, you should always use "Output (a)" or "Output (b)" to refer to the two outputs respectively.):
\end{tcolorbox}
\caption{Prompt for \texttt{Multi-role-round2} protocol described in \S\ref{sec:all_protocols}.}
\label{fig:prompt_multi-role_debate}
\end{figure*}

\begin{figure*}[t!]
\begin{tcolorbox}[colback=black!3!white, colframe=black!70!white, title=Multi-aspect-single, fontupper=\scriptsize, fonttitle=\footnotesize]
\textbf{[System Message]} \\
You are a helpful assistant who analyzes and evaluates the quality of two candidate outputs for a given instruction task based on a list of criteria, and makes a final decision on which output is better.
\newline

\textbf{[User Message]}\\
Given an instruction and two responses, Output (a) and Output (b), each aiming to fulfill the task, your task is to carefully analyze and evaluate each output based on a list of criteria.
\newline

Here are some general rules for the evaluation:

(1) You should prioritize evaluating whether the output honestly/precisely/closely executes the instruction, then consider its helpfulness, accuracy, level of detail, harmlessness, etc.

(2) Outputs should NOT contain more or less than what the instruction asks for, as such outputs do NOT precisely execute the instruction.
(3) You should avoid any potential bias and your judgment should be as objective as possible. For example, the order in which the outputs were presented should NOT affect your judgment, as Output (a) and Output (b) are equally likely to be the better one.
\newline

You should first provide an explanation of your evaluation, and then always end your response with either "Therefore, Output (a) is better." or "Therefore, Output (b) is better." verbatim.

Do NOT say both / neither are good.

Do NOT output any other words.

Do NOT say "Output (a) is better" or "Output (b) is better" at the beginning. You should do reasoning and thinking before claiming which output is better.
\newline

Here are some criteria to consider:

1. Text Quality: The response should be fluent, well-structured, and free of spelling and grammatical errors. It should also be coherent, with a clear and logical flow of ideas.

2. Information Richness: The response is encouraged to provide rich, detailed and professional information, e.g. by providing examples, explanations, citations, and additional information. This criterion is not applicable if the user asks for a short or direct answer without additional information.

3. User Intention Inference: If the user's intention is not clearly expressed by the query, the response should provide some relevant information, do some reasonable inference and ask more information for clarification. This criterion is not applicable if the user's intention is clearly expressed by the query.

4. Accuracy: All contents provided or mentioned in the response should be accurate and correct.

5. Completeness of Instruction Following: For all key instructions (e.g., answer multiple questions or perform multiple tasks) and explicit constraints (e.g. word count, response length limit, word usage, output format, etc.) provided by the user, the response should be complete in following all of them without any omission.
\newline

Consider how well each output meets the list of criteria and provide a comparative analysis.
After your analysis, make a final decision on which output is better overall. Provide a brief explanation of your evaluation, weighing the importance of each aspect, and make a final decision. Be mindful of the importance of each aspect in the context of the given instruction task, as some aspects may significantly influence the output's quality and relevance to the instruction, while others might be less critical.
\newline

Provide your response in the following format:

"""

1.Text Quality:

[Your analysis]

2. Information Richness:

[Your analysis]

3. User Intention Inference:

[Your analysis]

4. Accuracy:

[Your analysis]

5. Completeness of Instruction Following:

[Your analysis]
\newline

[Your overall evaluation and explanation, followed by the final decision]

"""
\newline
\# Instruction: \\
\{INSTRUCTION\}
\newline

\# Output (a):\\
\{OUTPUT\_1\}
\newline

\# Output (b):\\
\{OUTPUT\_2\}
\newline

\# Decision (Give an explanation of your evaluation followed by either "Therefore, Output (a) is better." or "Therefore, Output (b) is better." verbatim. Always claim which is better at the end. In your explanation, you should always use "Output (a)" or "Output (b)" to refer to the two outputs respectively.):
\end{tcolorbox}
\caption{Prompt for \texttt{multi-aspect-single} protocol described in \S\ref{sec:all_protocols}.}
\label{fig:prompt_multi_aspect_one}
\end{figure*}

\begin{figure*}[t!]
\begin{tcolorbox}[colback=black!3!white, colframe=black!70!white, title=Multi-aspect-two (aspect-wise analysis stage), fontupper=\footnotesize, fonttitle=\footnotesize]
\textbf{[System Message]} \\
You are a helpful assistant who analyzes and evaluates the quality of the outputs for a given instruction based on specific criteria.
\newline

\textbf{[User Message]}\\
Please provide a detailed analysis and evaluation of the two outputs based on the following criteria:

\{CRITERIA\_TEXT\}
\newline

Consider how well each output meets the criteria and provide a comparative analysis.
\newline

\# Instruction: \\
\{INSTRUCTION\}
\newline

\# Output (a): \\
\{OUTPUT\_1\}
\newline

\# Output (b): \\
\{OUTPUT\_2\}
\newline

\# Your analysis:
\end{tcolorbox}
\caption{Prompt for \texttt{multi-aspect-two} protocol described in \S\ref{sec:all_protocols}.  This is the prompt for aspect-wise analysis (the first stage) within the method.}
\label{fig:prompt_multi_aspect_two_analysis}
\end{figure*}

\begin{figure*}[t!]
\begin{tcolorbox}[colback=black!3!white, colframe=black!70!white, title=Multi-aspect-two (final evaluation stage), fontupper=\footnotesize, fonttitle=\footnotesize]
\textbf{[System Message]} \\
You are a helpful assistant who makes a final decision on which output is better based on the analysis of multiple aspects.
\newline

\textbf{[User Message]}\\
You have been provided with an instruction and two outputs, along with an analysis of each output based on several key aspects.

Your task is to carefully consider the analysis for each aspect and make a final decision on which output is better overall. Provide a brief explanation of your evaluation, weighing the importance of each aspect, and make a final decision. Be mindful of the importance of each aspect item in the context of the given instruction task, because some aspects may significantly influence the output's quality and relevance to the instruction, while others might be less critical.
\newline

Here are some rules of the evaluation:

(1) You should prioritize evaluating whether the output honestly/precisely/closely executes the instruction, then consider its helpfulness, accuracy, level of detail, harmlessness, etc.

(2) Outputs should NOT contain more/less than what the instruction asks for, as such outputs do NOT precisely execute the instruction.

(3) You should avoid any potential bias and your judgment should be as objective as possible. For example, the order in which the outputs were presented should NOT affect your judgment, as Output (a) and Output (b) are equally likely to be the better.
\newline

You should first provide an explanation of your evaluation, and then always end your response with either "Therefore, Output (a) is better." or "Therefore, Output (b) is better." verbatim.

Do NOT say both / neither are good.

Do NOT output any other words.

Do NOT say "Output (a) is better" or "Output (b) is better" at the beginning. You should do reasoning and thinking before claiming which is better.
\newline

\# Instruction: \\
\{INSTRUCTION\}
\newline

\# Output (a): \\
\{OUTPUT\_1\}
\newline

\# Output (b): \\
\{OUTPUT\_2\}
\newline

\# Here are the aspect-wise analyses provided by another helpful critic:

\{ANALYSIS\_HISTORY\}
\newline

\# Decision (Give a brief explanation of your evaluation followed by either "Therefore, Output (a) is better." or "Therefore, Output (b) is better." verbatim. Always claim which is better at the end. In your explanation, you should always use "Output (a)" or "Output (b)" to refer to the two outputs respectively.):
\end{tcolorbox}
\caption{Prompt for \texttt{multi-aspect-two} protocol described in \S\ref{sec:all_protocols}. This is the final evaluation prompt (the second stage) within the method.}
\label{fig:prompt_multi_aspect_two_final}
\end{figure*}

\begin{figure*}[t!]
\begin{tcolorbox}[colback=black!3!white, colframe=black!70!white, title=GPT4 Reference, fontupper=\footnotesize, fonttitle=\footnotesize]
\textbf{[System Message]} \\
You are a helpful assistant in evaluating the quality of the outputs for a given instruction. Your goal is to select the best output for the given instruction.
\newline

\textbf{[User Message]}\\
Select the Output (a) or Output (b) that is better for the given instruction. The two outputs are generated by two different AI chatbots respectively.
\newline

Here are some rules of the evaluation:
(1) You should prioritize evaluating whether the output honestly/precisely/closely executes the instruction, then consider its helpfulness, accuracy, level of detail, harmlessness, etc. \\
(2) Outputs should NOT contain more/less than what the instruction asks for, as such outputs do NOT precisely execute the instruction. \\
(3) You should avoid any potential bias and your judgment should be as objective as possible. For example, the order in which the outputs were presented should NOT affect your judgment, as Output (a) and Output (b) are **equally likely** to be the better. \\
\newline

Do NOT provide any explanation for your choice. \\
Do NOT say both / neither are good. \\ 
You should answer using ONLY "Output (a)" or "Output (b)". Do NOT output any other words. \\

\# Instruction: \\
\{INSTRUCTION\}
\newline

\# Output (a): \\
\{OUTPUT\_1\}
\newline

\# Output (b): \\
\{OUTPUT\_2\}
\newline

\# A reference output generated by a strong AI assistant:
\{REFERENCE\}
\newline

\# Which is better, Output (a) or Output (b)? Your response should be either "Output (a)" or "Output (b)":

\end{tcolorbox}
\caption{Prompt for \texttt{gpt4-reference} protocol described in \S\ref{sec:all_protocols}.}
\label{fig:prompt_gpt4_reference}
\end{figure*}

\begin{figure*}[t!]
\begin{tcolorbox}[colback=black!3!white, colframe=black!70!white, title=Prepair (pointwise analysis), fontupper=\footnotesize, fonttitle=\footnotesize]
\textbf{[System Message]} \\
You are a helpful assistant in evaluating the quality of the outputs for a given instruction. Your goal is to evaluate the quality of output for the given instruction.
\newline

\textbf{[User Message]}\\
Giving a brief explanation to evaluate the quality of the response to the given instruction. The output is generated by an AI chatbot.
\newline

Here are some rules of the evaluation:

(1) You should prioritize evaluating whether the output honestly/precisely/closely executes the instruction, then consider its helpfulness, accuracy, level of detail, harmlessness, etc.
\newline
(2) The model outputs should NOT contain more/less than what the instruction asks for, as such outputs do NOT precisely execute the instruction.
\newline
(3) You should avoid any potential bias and your judgment should be as objective as possible.
\newline

You should provide a brief explanation of your evaluation.
\newline
Your explanation should identify critical drawbacks in model outputs that do not meet the above evaluation rules.
\newline

\# Instruction: \\
\{INSTRUCTION\}
\newline

\# Output: \\
\{OUTPUT\}
\newline

\# Provide your explanation:
\end{tcolorbox}
\caption{Prompt for \texttt{prepair} protocol described in \S\ref{sec:all_protocols}.  This is the prompt for pointwise analysis (the first stage) within the protocol.}
\label{fig:prompt_prepair_pointwise}
\end{figure*}

\begin{figure*}[t!]
\begin{tcolorbox}[colback=black!3!white, colframe=black!70!white, title=Prepair (pairwise evaluation), fontupper=\footnotesize, fonttitle=\footnotesize]
\textbf{[System Message]} \\
You are a helpful assistant in evaluating the quality of the outputs for a given instruction. Your goal is to select the best output for the given instruction.
\newline

\textbf{[User Message]}\\
After giving a brief explanation, select the Output (a) or Output (b) that is better for the given instruction. The two outputs are generated by two different AI chatbots respectively.
\newline

Here are some rules of the evaluation:

(1) You should prioritize evaluating whether the output honestly/precisely/closely executes the instruction, then consider its helpfulness, accuracy, level of detail, harmlessness, etc.

(2) Outputs should NOT contain more/less than what the instruction asks for, as such outputs do NOT precisely execute the instruction.

(3) You should avoid any potential bias and your judgment should be as objective as possible. For example, the order in which the outputs were presented should NOT affect your judgment, as Output (a) and Output (b) are **equally likely** to be the
better.
\newline

You should first provide a brief explanation of your evaluation, and then always end your response with either "Therefore, Output (a) is better." or "Therefore, Output (b) is better." verbatim.

Do NOT say both / neither are good.

Do NOT output any other words.

Do NOT say "Output (a) is better" or "Output (b) is better" at the beginning.
\newline

You should do reasoning and thinking **before** claiming which is better. Your explanation should identify critical drawbacks in model outputs that do not meet the above evaluation rules.
\newline

\# Instruction: \\
\{INSTRUCTION\}
\newline

\# Output (a): \\
\{OUTPUT\_1\}
\newline

\# Output (b): \\
\{OUTPUT\_2\}
\newline

\# Here's the analysis for each output you wrote earlier: \\
\{PER OUTPUT ANALYSES\}
\newline

\# Your Response (Provide your evaluation and reasoning, followed by either "Therefore, Output (a) is better." or "Therefore, Output (b) is better." verbatim):

\end{tcolorbox}
\caption{Prompt for \texttt{prepair} protocol described in \S\ref{sec:all_protocols}. This is the pairwise evaluation stage (the second stage) within the method.}
\label{fig:prompt_prepair_pairwise}
\end{figure*}

\begin{table*}[ht]
  \centering
  \small
    \begin{tabular}{p{24.5em}c}
    \toprule
    \multicolumn{2}{l}{\textbf{Case 1}} \\
    \midrule
    \multicolumn{2}{l}{
    \parbox{15cm}{
    \textbf{Instruction:} Imagine that you are a guest speaker giving a lecture. Write a 1 to 2 sentences introduction about yourself. It should be generalizable to different fields.}
    }\\
    \\
    \multicolumn{1}{l}{\textbf{\red{Output (a)}}} & \multicolumn{1}{l}{\textbf{\lime{Output (b)} }} \\
    Hello everyone! My name is John Smith and I'm glad to be here today as a guest speaker. I'm a software engineer with over 10 years of experience ....
    & \multicolumn{1}{p{24.5em}}{Good evening everyone! My name is \_\_\_\_\_\_ and I'm excited to be here today to discuss the importance of \_\_\_\_\_\_.} \\
    \\
    \multicolumn{2}{l}{\textbf{Generation:}} \\
    \multicolumn{2}{p{50em}}{Output (a) provides a specific and concrete introduction about the speaker, including their name and profession, which is relevant to the lecture. On the other hand, Output (b) leaves blank spaces for the name and topic, making it unclear and incomplete. \red{Therefore, Output (a) is better.}} \\
    \\
    
    \multicolumn{2}{p{51em}}{\textbf{Comment:} Error type(s): \texttt{surface,overlook,counting}; \llama-3-70b failed to recognize that Output (a) violates both the length constraint and the generalization requirement specified in the instruction. Instead, it preferred Output (a) because it appeared more complete than Output (b).} \\
    \midrule
    \multicolumn{2}{l}{\textbf{Case 2}} \\
    \midrule
    \multicolumn{2}{p{51em}}{\textbf{Instruction:} "Determine the most common word in the text. Humans are created in the image of God, from a spiritual perspective and from a physical perspective."} \\
    \\
    \multicolumn{1}{l}{\textbf{\lime{Output (a)}}} & \multicolumn{1}{l}{\textbf{\red{Output (b)}}} \\
    "from", "a", and "perspective"
    & \multicolumn{1}{p{24.5em}}{The most common word in the text is "humans", which appears twice.} \\
    
    \multicolumn{2}{l}{\textbf{Generation:}} \\
    \multicolumn{2}{p{49em}}{The instruction asks to determine the most common word in the given text. Output (a) provides a list of words without specifying which one is the most common, whereas Output (b) clearly identifies the most common word as "humans" and provides the frequency of its appearance. Output (b) precisely executes the instruction by providing a single most common word, whereas Output (a) does not. \red{Therefore, Output (b) is better.}} \\
   \\
    \multicolumn{2}{p{50em}}{\textbf{Comment:} Error type(s): \texttt{surface,counting}; The model failed to identify the error in Output (b), where "humans" only appears once and is not the most common word. It preferred Output (b) due to its better formatting, despite the underlying error.} \\
    \midrule
    \multicolumn{2}{l}{\textbf{Case 3}} \\
    \midrule
    \multicolumn{2}{l}{\textbf{Instruction:} "Write a script for a 10-second conversation between two characters."} \\
    \\

    \multicolumn{1}{l}{\textbf{\red{Output (a)}}} & \multicolumn{1}{l}{\textbf{\lime{{Output (b)}}}} \\
    
    Person A: Hi, I'm John. It's nice to meet you. 
Person B: Hi John, I'm Jane. It's nice to meet you too.
.... (7 more turns)
    & \multicolumn{1}{p{24.5em}}{
Person A: Hi, how are you?
Person B: I'm doing well, how about you? 
(1 more turn)
    } \\
    \\

    \multicolumn{2}{l}{\textbf{Generation:}} \\
    \multicolumn{2}{p{49em}}{The instruction asks for a script for a 10-second conversation between two characters. Output (a) provides a conversation that is too long and detailed for a 10-second conversation, as it includes multiple topics and exchanges. Output (b), on the other hand, provides a brief and concise conversation that can fit within a 10-second time frame. However, Output (b) does not provide a complete conversation, as it only includes two exchanges and does not provide a clear conclusion or ending.

\red{Therefore, Output (a) is better.}} \\
\\

    \multicolumn{2}{p{50em}}{\textbf{Comment:} Error type(s):  \texttt{surface,overlook}; \llama-3-70b overlooked a clear violation of the instruction in Output (a), where the conversation exceeds the 10-second limit. Despite this, it preferred the output as the more "complete" conversation, even though completeness was not a required attribute in the instruction.} \\
    \bottomrule
    \end{tabular}%
  \caption{Case study (Appendix~\S\ref{appx_case_study}) showing error patterns in \llama-3-70b generations under \texttt{cot} protocol (\S\ref{sec:all_protocols}).}
  \label{tab:appx_error_case}
\end{table*}

\clearpage


\end{document}